\documentclass[pmlr]{jmlr}

\RequirePackage{graphicx}
\usepackage{booktabs}
\usepackage{longtable}
\usepackage{multirow}
\usepackage{graphicx}
\usepackage{array}
\usepackage{adjustbox}
\usepackage{array}
\usepackage[table]{xcolor}
\usepackage{enumitem}
\usepackage[load-configurations=version-1]{siunitx}
\usepackage{float}
\makeatletter
\def\set@curr@file#1{\def\@curr@file{#1}}
\makeatother

\makeatletter
\def\@titlefoot{}
\def\ps@jmlrtps{%
  \let\@mkboth\@gobbletwo
  \def\@oddhead{}%
  \let\@evenhead\@oddhead
  \def\@oddfoot{}%
  \let\@evenfoot\@oddfoot
}
\makeatother

\title[Hospitalization Forecasting Evaluations]{Context-Aware Hospitalization Forecasting Evaluations for Decision Support using LLMs}

\author{\Name{Rhea Makkuni} \and \Name{Ananya Joshi}\\
\addr Johns Hopkins University\\ Baltimore, USA}

\begin{document}
\maketitle

\begin{abstract}
Medical and public health experts must make real-time resource decisions, such as expanding hospital bed capacity, based on projected hospitalization trends during large-scale healthcare disruptions (e.g., operational failures or pandemics). Forecasting models can assist in this task by analyzing large volumes of resource-related data at the facility level, but they must be reliable for decision-making under real-world data conditions. Recent work shows that large language models (LLMs) can incorporate richer forms of context into numerical forecasting. Whereas traditional models rely primarily on temporal context (i.e., past observations), LLMs can also leverage non-temporal public health context such as demographic, geographic, and population-level features. However, it remains unclear how these models should be used to produce stable or decision-relevant predictions in real-world healthcare settings. To evaluate how LLMs can be effectively used in this setting, we evaluate three approaches across 60 counties with low-,mid-, and high- hospitalization intensities in the United States: direct LLM-based forecasting, classical time-series models, and a context-augmented hybrid pipeline(\textsc{HybridARX}) that incorporates LLM-derived signals into structured models.   Because the goal is operational decision-making rather than error minimization alone, we evaluate performance with bias and lead--lag alignment in addition to standard forecasting metrics. Our results show that \textsc{HybridARX} improves over classical ARX by yielding more stable and better-calibrated forecasts, particularly when incorporating noisy contextual signals into structured time-series models. These findings suggest that, in non-stationary healthcare resource forecasting, LLMs are most useful when embedded within structured hybrid models.\end{abstract}

\section{Introduction}
Accurate hospitalization forecasting is vital to healthcare systems because it enables timely resource allocation, staffing, and capacity planning to maintain quality of care during periods of fluctuating demand \citep{preiss2022,sandhu2022}. As demonstrated by the COVID-19 surges from 2020--2023, unexpected demand can overwhelm the ability of healthcare systems to allocate resources in real time, leading to strained intensive care units \citep{sandhu2022} and widespread disruption of essential services such as chemotherapy and rehabilitation \citep{berger2022}. As a result, healthcare systems, including hospitals and public health agencies, require reliable decision support to both (1) guide resource allocation during large-scale healthcare disruptions and (2) respond effectively to emerging system strain \citep{moynihan2021,kadri2021}. 

However, there are practical challenges within the healthcare setting, such as the facility-level heterogeneity in system capacity, including variation in staffing, bed availability, and equipment. Systems with limited capacity, particularly those serving socially vulnerable or rural populations \citep{tsai2022}, are especially sensitive to local forecasting errors, where even small deviations can lead to substantial misallocation. This means that models that perform well in aggregate may still fail in practice if they do not capture regional variation, thereby compounding disparities between resource-constrained and well-resourced hospitals. 
\begin{figure*}[t]
\includegraphics[width=1.05\textwidth]{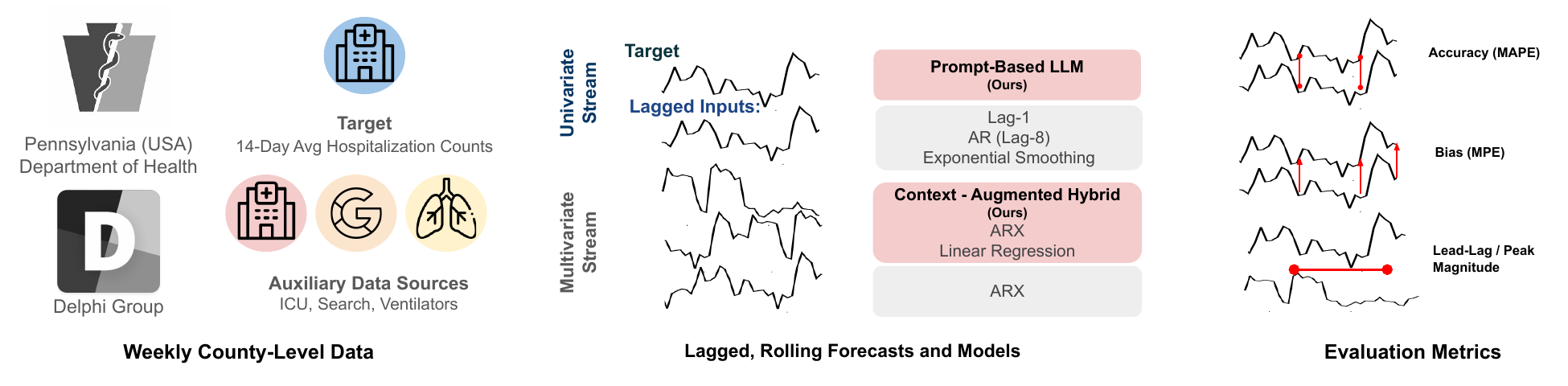}
\caption{Evaluation pipeline to compare metrics from LLM-based methods. Data is pulled from multiple data sources with varying granularity and are processed to be standardized input. We compare multiple forecasting algorithms on metrics relevant to decision-making tasks. }
\label{fig:methods_overview}
\end{figure*}
Accordingly, many general forecasting models that are used as the backbone for resource allocation algorithms do not reliably support decision-making. Often, Lag-1 models, or simply reusing the previous measurement as the forecast,  have the highest accuracy values. However,  these are not necessarily meaningful for decision-making. Other models, such as autoregressive (AR) models, rely heavily on recent observations and are therefore unable to anticipate regime shifts or rapid changes in disease dynamics under non-stationary conditions \citep{cramer2022evaluation}. This effect is pronounced across different population levels, where the operational cost of forecasting error can vary substantially, which motivates the need for methods that can incorporate additional non-temporal contextual information. 

Despite their advanced contextual reasoning capabilities,  it remains unclear how Large Language Models (LLMs) can be applied to produce stable decision making forecasts in real world healthcare settings. While LLMs can encode and transform heterogeneous contextual signals, they are prone to hallucination and instability. Thus, rather than examining LLMs as standalone forecasters, we examine whether they are more effective when used to generate contextual signals that can be incorporated into structured classical forecasting pipelines. In this work, we systematically evaluate  three approaches using county-level COVID-19 hospitalization data from the Pennsylvania Department of Health, which provides a consistent setting across counties with varying hospitalization intensities, to understand how LLMs can be effectively integrated into healthcare forecasting pipelines. 

\begin{enumerate}
\item Direct LLM prediction with contextual and temporal information embedded in prompts
\item \textsc{HybridARX}:  A two-stage hybrid approach in which an LLM first predicts next-week values of leading indicators (i.e., observable signals that tend to precede hospitalizations), and these LLM-generated indicator forecasts are then supplied as exogenous inputs to classical time-series models for the final hospitalization forecast.
\item Classical time-series baselines, including univariate and autoregressive models (Lag-1, AR(1), exponential smoothing), as well as multivariate autoregressive models with exogenous inputs (ARX)
\end{enumerate}

We conduct a county-level analysis across 60 counties within a single state, providing a shared reporting and policy environment that enables controlled comparison across heterogeneous low-, mid-, and high-intensity regions (defined using tertiles of mean weekly hospitalizations from Pennsylvania Department of Health data). Our results show that prompt-only LLM forecasts capture broad hospitalization trends but do not consistently outperform simple statistical baselines. In contrast, \textsc{HybridARX} improves over classical ARX by reducing large errors in several settings and producing more stable forecasts when noisy contextual signals are incorporated into a structured time-series model, although it does not uniformly outperform simpler methods. Instead, \textsc{HybridARX} demonstrates that LLMs are valuable as contextual signal encoders that transform heterogeneous information into structured predictors, which can then be incorporated into traditional time-series forecasting pipelines.

{\subsection{Generalizable Insights about Machine Learning in the Context of Healthcare}

    In non-stationary and noisy healthcare settings, simply adding non-temporal contextual information does not necessarily improve forecasting. Instead, what matters is whether these contextual signals are incorporated within structured forecasting pipelines. Our results suggest that LLMs are most useful not as standalone forecasters, but as components within hybrid pipelines that transform heterogeneous contextual inputs into structured predictors for classical time-series models. In particular, using LLM outputs as inputs to forecasting pipelines can improve the stability and consistency of outputs.  This distinction is especially important in healthcare settings, where decision-making has low tolerance for systematic bias or delayed signals, and relying on unconstrained black-box forecasts can lead to unstable or poorly calibrated predictions that negatively impact resource allocation. Instead, incorporating LLMs allows contextual information to be leveraged while maintaining stability and interpretability in high-stakes decision environments, where even small forecast errors can affect the quality of patient care, staff workload, and resource utilization.

\section{Evaluation Constraints for Healthcare Forecasting}
For hospitalization forecasting algorithms to support healthcare resource allocation in practice, they must satisfy three design goals: they must remain reliable under non-stationary and noisy data,  incorporate local context without destabilizing forecasts, and be evaluated using decision-relevant criteria. \\

\noindent{\textbf{1. Leading indicators are unstable and historical relationships drift:}} 
Forecasting-relevant leading indicators, such as disease transmission rates or measurement quality, can evolve rapidly over short time horizons, which means that the relevance of historical data may quickly diminish \citep{cramer2022evaluation, joshi2024outlier}. This may help explain why simple baseline approaches, such as lag-1 predictors, are often among the best-performing algorithms at the weekly scale, but are not as informative for resource allocation over time \citep{cramer2022evaluation,hyndman2018}. The data types in this setting contribute to why general forecasting algorithms remain insufficient. For example, resource data provide only a partial, time-lagged view of disease transmission and are often affected by data irregularities and reporting anomalies \citep{reinhart2021open,mcdonald2021}. This motivates the need for approaches that remain stable under short rolling windows, particularly when signal sparsity and noise are pronounced due to reporting delays and measurement variability.\\

\noindent\textbf{2. Context matters and must be treated carefully:} 
Healthcare systems operate under continuously evolving policies and are shaped by regional contexts. Effective decision-making depends on regional and temporal forecasts rather than previous temporal context alone. Decision-makers need contextualized model outputs for operational planning within specific regions \citep{cramer2022evaluation}, as misinformed local allocation decisions can overextend providers and compromise care delivery \citep{sandhu2022}. \\

\noindent\textbf{3. Decision-relevant Evaluation:} 
Healthcare experts must make real-time decisions, as acquiring and deploying resources is costly and time-consuming, especially during surges \citep{sandhu2022,cramer2022evaluation}. Thus, evaluation criteria must extend beyond point accuracy to include metrics relevant to decision-making under uncertainty \citep{bracher2021}, such as calibrated uncertainty estimates. In addition, evaluation settings must account for real-world disruptions. Models that systematically underpredict demand may fail to capture early warning signals; conversely, models that overreact can lead to inconsistent decision-making that requires repeated and unnecessary resource adjustments. Therefore, evaluation must account for systematic bias (consistent over- or under-prediction) and temporal stability, as model performance can vary over time in non-stationary healthcare settings \citep{cramer2022evaluation}. This variability can undermine reliable operational decision-making, even when average accuracy is acceptable.

\section{Related Work}

Real-time hospitalization forecasting is essential for decision-making, enabling hospitals to maintain high-quality care, especially when resources such as beds are limited \citep{preiss2022}. In these settings, forecasting utility depends not only on aggregate accuracy but also on calibration and reliability, which directly affect planning.

Prior work has applied machine learning methods such as random forests, gradient-boosted trees (XGBoost), support vector machines (SVM), and elastic nets to forecast hospital admissions \citep{alaawar2025}. Time-series models such as exponential smoothing state space models (ETS), TBATS, and additive models with nonlinear trends and seasonal effects (Prophet) have been benchmarked against these approaches in disrupted settings, such as psychiatric admissions during the pandemic \citep{wolff2022}. More broadly, large-scale epidemiological forecasting efforts often include hospital resource targets such as bed occupancy \citep{mellor2025}. CDC forecasting efforts, such as FluSight \citep{cdc_flusight} and the COVID-19 Forecast Hub \citep{reich2022covidhub}, include multiple models, which are generally trend-extension and mechanistic hybrid approaches. Variants of these methods, such as ARIMAX or mechanistic models (e.g., SEIR-based models), also exhibit high accuracy \citep{somyanonthanakul2022}. Additionally, prior work has incorporated non-temporal context through coarse, bucketed representations, such as stratifying populations by age, geography, or risk group, or including aggregated demographic covariates in regression-based models. Newer foundation time-series models, such as MOMENT, extend this idea by training transformers on large collections of time-series data \citep{goswami2024}, including epidemiological data, but require a large number of historical samples (e.g. 512) that may not be reliable or relevant in a nonstationary setting, where there may be between 8-10 reliable previous measurements. 

While promising, many of these approaches were evaluated solely from an accuracy perspective instead of a decision-making lens, masking challenges with real world data forecasting. First, as a result of the structure of the models, they still rely on stable historical relationships and require frequent refitting. Second, many leading indicators can be inconsistent because their conversion rates (the probability that a leading indicator results in hospitalization) are influenced by contextual factors, making resource allocation during surges particularly challenging \citep{reich2023leadingindicator}. In addition, any bucket-based approaches that treat additional context as fixed, discretized inputs may disproportionately impact regions with less representative buckets (e.g. lower hospitalization intensity). Many such methods also rely on expert-selected clinical characteristics, hand-crafted preprocessing, and tuned hyperparameters, and they rarely operate directly on unstructured sources or without significant maintenance \citep{baik2022}. This requirement may be incompatible with the budgets and real-time constraints faced by many public health departments. As a result, resource allocation decisions frequently rely on ad hoc adjustments or expert judgment layered on top of trend-based forecasts, reducing reproducibility and scalability across regions and time \citep{cramer2022evaluation}

LLMs offer a different form of flexibility. LLMs have been explored for forecasting the short-term spread of disease outbreaks by reformulating real-time forecasting as a text reasoning problem \citep{du2024}, as well as for monitoring time-series models to flag implausible or inaccurate predictions in large-scale retail settings \citep{bhan2025forecastcritic}. Despite these advances, the use of LLMs for operational hospital forecasting and, more generally, as primary forecast generators remains largely unexplored. 

This work addresses this gap by systematically comparing established time-series baselines, direct LLM-based forecasting, and a context-augmented hybrid pipeline (\textsc{ HybridARX}) using county-level COVID-19 hospitalization data. It provides a controlled, county-level evaluation of LLM-based, classical, and hybrid forecasting strategies for short-horizon hospitalization prediction, isolating when contextual information improves operational reliability.

\section{Methods}

For each county, we consider a weekly time-series of COVID-19 hospitalizations. Let $y_t$ denote the observed hospitalization level at week $t$, and let $z_t$ denote a vector of auxiliary contextual indicators. The forecasting task is to produce a one-week-ahead prediction:

\[
\hat{y}_{t+1} = f(y_{t-L:t}, z_{t-L:t})
\]

where $f$ is a forecasting model that could be either statistical, LLM-based, or hybrid. All models are evaluated using a rolling-origin one-step-ahead forecasting protocol to prevent data leakage.

We consider three forecasting model classes: (1) a prompt-only LLM approach that generates predictions using recent observations expressed as structured text; (2) classical statistical baselines, including the lag-1 model, autoregressive (AR) models, and exponential smoothing, which serve as interpretable reference methods without explicit contextual inputs; and (3) the \textsc{HybridARX} approach, which combines LLM-derived contextual signals with classical time-series models to improve stability, calibration, and temporal alignment in forecasting.

For each county and forecast time $t+1$, the model is provided with the most recent $L=8$ weekly observations of the hospitalization series.

\subsection{Data Selection and Extraction}

We constructed a weekly panel using the following public data sources:

\medskip
\noindent\textit{Hospitalization and hospital capacity data:} Obtained from the Pennsylvania Department of Health~\citep{pa_covid_hospitalizations}.\footnote{Seven counties (Forest, Juniata, Perry, Pike, Snyder, Cameron, and Sullivan) were excluded from per-county evaluation due to data missingness and reporting irregularities across the study period.} The dataset includes weekly reported values derived from daily data, including 14-day averaged daily counts of COVID-19 hospitalizations, as well as adult and pediatric ICU and ward bed availability and occupancy. These reported counts may be affected by testing practices, reporting delays, backlogs, and routine data cleaning. County assignment was based on patient or provider address and may be imperfect for ZIP codes that cross county boundaries.

\medskip
\noindent\textit{Leading indicators:} We used the Delphi COVIDcast API~\citep{delphi2020covidcast} to obtain leading indicators, including Facebook symptom survey measures (mask wearing, vaccine acceptance, and COVID-like symptoms in individuals and in the community), Google symptom search intensity for anosmia and ageusia, and COVID-related outpatient and emergency department visit rates, which are reported at mixed geographic resolutions. We aggregated daily values to weekly sums. When signals were unavailable at the county level for a given week, the corresponding Pennsylvania state-level signal was used.

We also used mobility indicators derived from Google COVID-19 Community Mobility Reports~\citep{google_covid_mobility}, based on percentage changes in visits to retail and recreation and residential locations. These categories were selected because they reflect high-contact public activity and voluntary risk avoidance.

\medskip
\noindent\textit{Population estimates:} Population estimates from the 2020 U.S. Census were used to stratify counties~\citep{uscensus_pa_profile}. Counties were ranked by their mean weekly hospitalization counts over March~2020--January~2022 and categorized into lower, middle, and upper tertiles. The final panel includes these features ($x$) and the target hospitalization outcome ($y$), defined as the 14-day average number of hospitalized COVID-19 patients.

\subsection{Leading Indicators and Correlation Analysis}

\begin{figure*}[t]
    \centering
    \includegraphics[width=\textwidth]{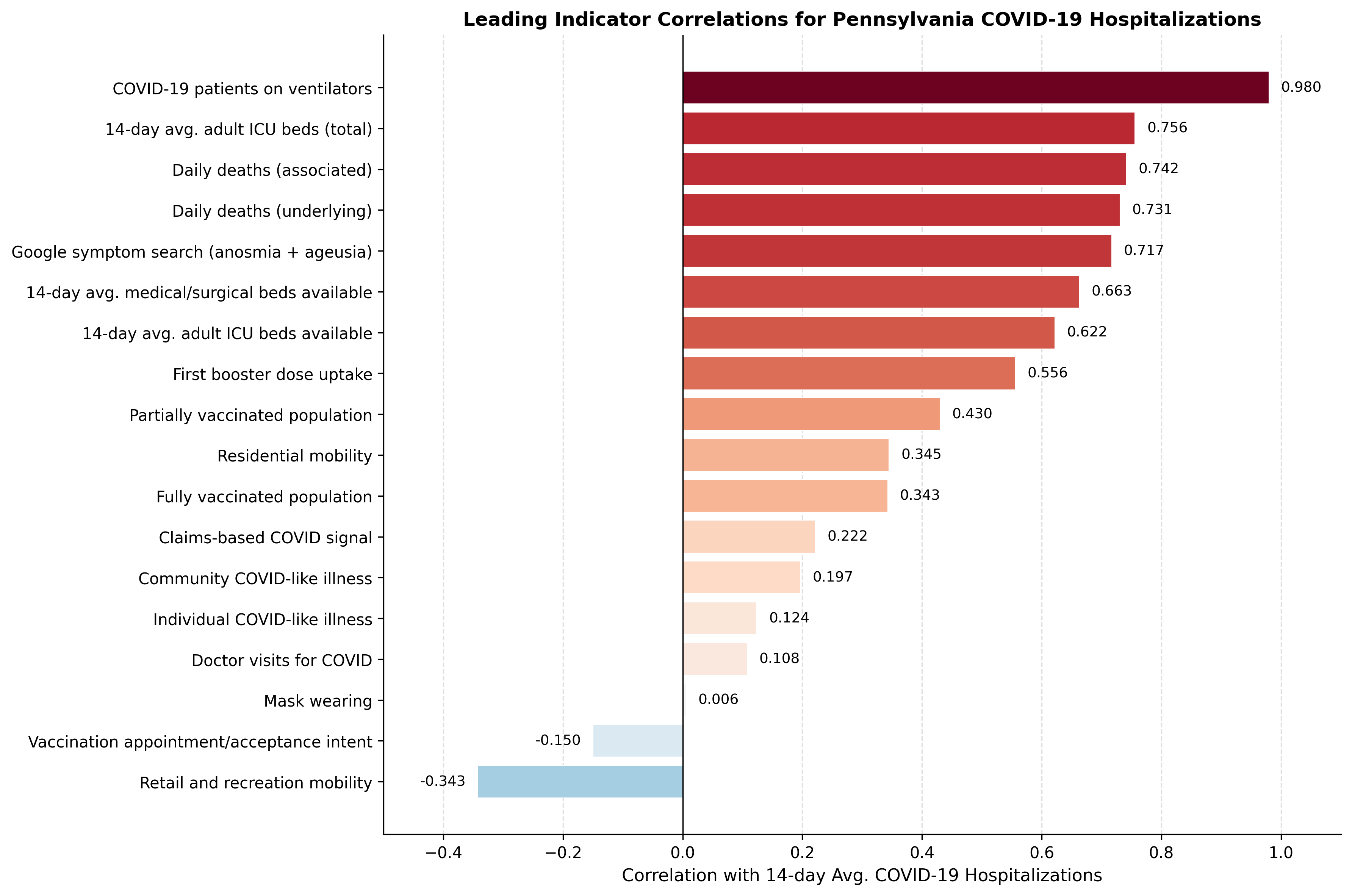}
    \caption{Pearson correlation coefficients between candidate leading indicators and county-level 14-day average COVID-19 hospitalizations. Hospital capacity measures and anosmia/ageusia search volume exhibit the strongest positive associations.}
    \label{fig:indicator-correlations}
\end{figure*}
To identify the most informative leading indicators, we analyzed associations between hospitalization outcomes and a broad set of candidate signals from the panel data. Consistent with the forecasting setup, all indicators were aligned to a weekly resolution, with hospitalization outcomes defined as the 14-day average number of hospitalized COVID-19 patients reported each week. 

We calculated Pearson correlation coefficients between each indicator and the hospitalization outcome using only weeks in which both variables were observed. Indicators were ranked by the magnitude of their correlations to assess their potential relevance as leading signals.  The resulting coefficients, shown in Fig.~\ref{fig:indicator-correlations}, highlight strong associations between hospitalizations and measures of hospital capacity (e.g., ventilator utilization and ICU beds), as well as anosmia/ageusia search activity, and are used to guide our evaluation of how contextual signals impact forecasting performance.

\subsection{Prompt-Only LLM [Univariate]}

In the prompt-only LLM setting, the model is used as a univariate forecaster. Each prompt consists of (1) the county identifier, (2) a chronologically ordered list of the previous eight weeks of numeric hospitalization values, where each value corresponds to the reported 14-day average number of hospitalized COVID-19 patients for that week, and (3) an instruction to return a single numerical prediction for the subsequent week. The model output is then parsed to a numeric value corresponding to $y$.

\subsection{Classical Time-Series Baselines [Univariate and Multivariate]}

We evaluate several classical statistical baselines, each fitted independently for each county using a rolling window of $L = 8$ weekly observations.

\begin{itemize}
    \item \textit{Lag-1 baseline}: A naïve persistence model predicting the next week’s hospitalization level as the most recently observed value.
    \item \textit{Autoregressive model (AR(1))}: A first-order autoregressive model with an intercept, trained on the previous eight weeks of hospitalization data.
    \item \textit{Exponential smoothing}: An additive exponential smoothing model with trend is used. Seasonal components are omitted because our evaluation provides insufficient historical context to estimate seasonality.
    \item \textit{Autoregressive model with exogenous inputs (ARX)}: An AR(1) model augmented with weekly exogenous variables, including adult ICU beds ($X_B$), ventilator utilization ($X_V$), and anosmia/ageusia search volume ($s_t$).
\end{itemize}

\subsection{Context-Augmented Hybrid Model [Multivariate] (\textsc{HybridARX})}
\label{sec:context_augmented_hybrid}

To examine how LLM-derived contextual signals affect forecasting performance, we construct a context-augmented hybrid pipeline  (\textsc{HybridARX}) that separates context generation from final prediction.  
In the first stage, a LLM is used to predict next-week contextual variables:
\begin{itemize}
    \item $\hat{X}_{B,t+1}$: adult ICU beds,
    \item $\hat{X}_{V,t+1}$: ventilator utilization,
    \item $\hat{s}_{t+1}$: anosmia/ageusia search volume
\end{itemize}

In the second stage, only the LLM-predicted contextual variables
\[
(\hat{X}_{B,t+1}, \hat{X}_{V,t+1}, \hat{s}_{t+1})
\]
are supplied to classical forecasting models (ARX and linear regression) to produce the final hospitalization forecast $\hat{y}_{t+1}$. We consider both ARX and linear regression to test the impact of temporal dependence assumptions. Additional implementation details, including prompt templates, preprocessing steps, and hyperparameter settings, are provided in Appendices A and B.

\section{Evaluation Formulation}
\label{sec:evaluation}

Operational decision-making depends not only on accuracy but also on timing, stability, and systematic bias. We evaluate model performance using accuracy, bias, and lead--lag behavior (stratified by hospitalization intensity) to capture error magnitude, direction, temporal alignment, and run-to-run variability in LLM-based forecasts. 

All methods are evaluated under a consistent rolling-origin, one-step-ahead forecasting protocol at a weekly resolution, using the most recent $L=8$ observations at each forecast time. To quantify variability in LLM-based methods, the full rolling-window forecasting procedure is repeated $n=3$ times to study model variability. For each county--week--model combination, we record the predicted hospitalization level and the corresponding percent error for every run, using the following performance metrics:

\noindent\textit{1. MAPE Analysis.} Forecasting accuracy is assessed using mean absolute percent error (MAPE), defined as
\[
\text{MAPE} = 100 \times \frac{\lvert \hat{y}_{t+1} - y_{t+1} \rvert}{\max(y_{t+1}, 1)},
\]
where $y_{t+1}$ denotes the observed 14-day average hospitalization level in the subsequent week, and $\hat{y}_{t+1}$ denotes the corresponding forecast. MAPE is computed at each forecast point and then averaged across counties and evaluation weeks.\\

\noindent\textit{2. MPE Analysis (Bias).} To quantify systematic over- or under-prediction, we compute mean percent error (MPE):
\[
\mathrm{MPE}(t)
=
100 \times
\frac{\hat{y}_{t+1} - y_{t+1}}
{\max(y_{t+1},1)}.
\]
Positive values indicate systematic overestimation, while negative values indicate systematic underestimation. \\

\noindent\textit{3. Lead--Lag Analysis.} In addition to pointwise error, we evaluate temporal alignment between predicted and realized hospitalization trajectories. For each county, we compute differences 
,$ \Delta y_t = y_t - y_{t-1}$ and $\Delta \hat{y}_t = \hat{y}_t - \hat{y}_{t-1} $ , and define a lagged trend correlation for integer offsets $\ell \in [-\ell_{\max}, \ell_{\max}]$: $\rho(\ell)
=
\mathrm{corr}\!\left(
\Delta \hat{y}_{t-\ell},\,
\Delta y_t
\right)$, 
computed over all overlapping evaluation weeks. We then define the lead--lag estimate as
\[
\ell^\star
=
\arg\max_{\ell \in [-\ell_{\max},\ell_{\max}]}
\rho(\ell).
\]
Positive values of $\ell^\star$ indicate that the forecasted trend leads the realized trend (i.e., anticipates changes), while negative values indicate lagging behavior. We report the mean and standard deviation of $\ell^\star$ across counties. We additionally report the peak correlation value $\rho^\star
=
\max_{\ell \in \left[-\ell_{\max},\, \ell_{\max}\right]}
\rho(\ell)$, which quantifies the strength of trend alignment regardless of whether the model leads or lags. For this specific metric, we omit the Lag-1 baseline from lead--lag evaluation because, under
$\hat{y}_t = y_{t-1}$, we have $\Delta \hat{y}_t = \Delta y_{t-1}$. This makes perfect alignment at $\ell=-1$ (and thus $\rho^\star=1$) mathematically guaranteed.

\section{Results}

\begin{table}[h]
\centering
\begin{tabular}{lccc}
\toprule
 & \multicolumn{3}{c}{\textbf{MAPE:} Hospitalization Intensity (Lower is Better) $\downarrow$} \\
\cmidrule(lr){2-4}
\textbf{Model} & \textbf{Low} & \textbf{Mid} & \textbf{High} \\
\midrule

\multicolumn{4}{l}{\textit{Classical Baselines}} \\
\addlinespace[2pt]
Lag-1                 & \textbf{23.80 $\pm$ 3.97} & \textit{22.66 $\pm$ 3.49} & \textit{17.85 $\pm$ 2.68} \\
AR(1)                 & 32.44 $\pm$ 7.66 & \textit{26.93 $\pm$ 5.83} & \textit{20.10 $\pm$ 3.87} \\
Exp.\ Smoothing       & \textit{26.47 $\pm$ 4.94} & \textbf{22.54 $\pm$ 4.44} & \textbf{15.49 $\pm$ 4.17} \\
ARX                   & 35.74 $\pm$ 7.75 & 31.17 $\pm$ 7.13 & 26.06 $\pm$ 14.91 \\
\midrule

\multicolumn{4}{l}{\textit{LLM-Based}} \\
LLM (Prompt-Only)     & \textit{25.44 $\pm$ 4.15} & \textit{22.59 $\pm$ 4.18} & \textit{19.59 $\pm$ 3.30} \\
\midrule

\multicolumn{4}{l}{\textit{Context-Augmented Hybrid}} \\
Hybrid ARX            & 31.16 $\pm$ 7.20 & \textit{24.94 $\pm$ 5.52} & \textit{18.84 $\pm$ 3.87} \\
Hybrid Linear Reg.    & 39.91 $\pm$ 9.82 & 38.15 $\pm$ 10.31 & 25.76 $\pm$ 6.71 \\
\bottomrule
\end{tabular}
\caption{MAPE (\%) by forecasting model across hospitalization-intensity tertiles (mean $\pm$ SD across counties). Bold indicates the best mean value in each column. Italics indicate values with overlapping intervals relative to the best mean in that column. Prompt-based LLMs consistently perform well on this traditional forecasting metric, likely because simple trend-extension techniques also perform well under this metric.}
\label{tab:mape_intensity}
\end{table}
\subsection{Accuracy: MAPE}

Across all models,  MAPE (Tab \ref{tab:mape_intensity}) is highest in low-intensity counties and lowest in high-intensity counties, supporting that low-volume counties are harder to forecast because small absolute fluctuations produce disproportionately large relative errors. Among the classical baselines, exponential smoothing and the Lag-1 model perform competitively in high-intensity counties. In contrast, AR models show substantially higher variance, particularly when exogenous inputs are included. Classical ARX performs poorly in high-intensity counties and mid-intensity counties, which suggests that noise in the exogenous variables destabilized the regression and negatively impacted forecasting accuracy. The prompt-only LLM shows relatively consistent performance across county types and is competitive with exponential smoothing , which once again supports that, even without exogenous variables, LLMs can capture short-term temporal structure and smooth random fluctuations in hospitalization trends.

 The \textsc{HybridARX} model improves substantially on ARX across the board, and is competitive in both mid-intensity and high-intensity counties. This suggests that contextual signals do not automatically improve forecasting performance. Unlike ARX, which incorporates temporal dependence through past hospitalization values, linear regression relies solely on contextual predictors and therefore cannot capture the temporal relationships present in hospitalization data. Therefore, the additional signals introduce noise and reduce forecasting accuracy. This is corroborated by the poor performance exhibited by Hybrid Linear Regression across all county types.

\subsection{Directional Bias (MPE)}
\begin{table}[t]
\centering
\begin{tabular}{lcccc}
\toprule
 & \multicolumn{4}{c}{\textbf{MPE:} Hospitalization Intensity} \\
\cmidrule(lr){2-5}
\textbf{Model} & \textbf{Overall} & \textbf{Low} & \textbf{Mid} & \textbf{High} \\
\midrule

\multicolumn{5}{l}{\textit{Classical Baselines}} \\
\addlinespace[2pt]
Lag-1              & \textit{+0.7 $\pm$ 1.5}  & \textit{+1.1 $\pm$ 1.4}  & \textbf{+0.4 $\pm$ 1.9}  & \textbf{+0.6 $\pm$ 1.2}  \\
AR(1)              & \textit{+3.2 $\pm$ 6.5}  & +9.5 $\pm$ 6.5  & \textit{+0.9 $\pm$ 4.1}  & \textit{-0.9 $\pm$ 2.6}  \\
Exp.\ Smoothing    & \textit{-3.4 $\pm$ 3.5}  & \textbf{-0.6 $\pm$ 3.7}  & -4.4 $\pm$ 2.7  & -5.4 $\pm$ 1.9  \\
ARX                & +5.5 $\pm$ 10.8 & +8.6 $\pm$ 8.1  & +3.6 $\pm$ 5.9  & +4.1 $\pm$ 15.7 \\
\midrule

\multicolumn{5}{l}{\textit{LLM-Based}} \\
LLM (Prompt-Only)  & \textbf{+0.6 $\pm$ 3.1}  & \textit{+2.5 $\pm$ 2.3}  & \textit{+1.0 $\pm$ 3.0}  & \textit{-1.7 $\pm$ 2.4}  \\
\midrule

\multicolumn{5}{l}{\textit{Context-Augmented Hybrid}} \\
Hybrid ARX         & \textit{+0.7 $\pm$ 6.1}  & +5.2 $\pm$ 7.7  & \textit{-0.4 $\pm$ 4.3}  & -2.7 $\pm$ 2.1  \\
Hybrid Linear Reg. & +7.5 $\pm$ 8.4  & +8.3 $\pm$ 8.8  & +10.7 $\pm$ 9.3 & +3.4 $\pm$ 5.4  \\
\bottomrule
\end{tabular}

\caption{Mean percent error (MPE, \%) by forecasting model across hospitalization-intensity strata (mean $\pm$ SD across counties). Bold indicates the value closest to zero in each column. Italics indicate values with overlapping intervals relative to the best-performing model in that column. Positive values indicate overprediction, while negative values indicate underprediction.}

\label{tab:mpe_intensity}
\end{table}
Mean Percent Error (MPE) captures systematic over- and under-prediction through positive and negative values, respectively (Tab \ref{tab:mpe_intensity}). In most cases, lag-1 was the most calibrated model, with MPE very close to zero across all hospitalization intensities. Interestingly, the prompt-only LLM model also exhibited bias close to zero overall ($+0.6 \pm 3.1$), with relatively small deviations across hospitalization intensities. Once again, \textsc{HybridARX} performs better than ARX, supporting that they may improve calibration  and reduce the uncertainty when used as part of hybrid pipelines in decision-making models. 

\subsection{Lead--Lag Analysis}

\begin{table}[h]
\centering
\begin{adjustbox}{max width=0.95\columnwidth}
\begin{tabular}{lcccc}
\toprule
 & \multicolumn{4}{c}{\textbf{Lead--Lag Analysis}} \\
\cmidrule(lr){2-5}
\textbf{Model} 
& \textbf{$\ell^\star$ Mean} 
& \textbf{$\ell^\star$ SD} 
& \textbf{$\rho^\star$ Mean} 
& \textbf{$\rho^\star$ SD} \\
\midrule

\multicolumn{5}{l}{\textit{Classical Baselines}} \\
\addlinespace[2pt]
AR(1)                 & -0.97 & 0.26 & 0.802 & 0.139 \\
ARX                   & -0.84 & 0.76 & 0.563 & 0.133 \\
Exp.\ Smoothing       & -1.00 & 0.00 & \textbf{0.872} & \textbf{0.036} \\
\midrule

\multicolumn{5}{l}{\textit{LLM-Based}} \\
LLM (Direct)          & -1.00 & 0.00 & \textit{0.859} & \textit{0.057} \\
\midrule

\multicolumn{5}{l}{\textit{Context-Augmented Hybrid}} \\
Hybrid ARX            & -0.97 & 0.18 & 0.771 & 0.123 \\
Hybrid Linear Reg.    & -1.21 & 0.45 & 0.683 & 0.124 \\
\bottomrule
\end{tabular}
\end{adjustbox}

\caption{Lead--lag summary across forecasting models (mean $\pm$ SD across counties). Bold indicates the highest trend alignment ($\rho^\star$). Italics indicate values with overlapping intervals relative to the best-performing model.}

\label{tab:lead_lag_summary}
\end{table}

Because most methods yield mean lead--lag estimates close to $\ell^\star = -1$, we primarily focus on $\rho^\star$, the peak trend-correlation metric, when assessing temporal alignment. Exponential smoothing and prompt-only LLM forecasts show the strongest trend alignment, while AR(1) and ARX achieve moderate and weaker alignment. Hybrid ARX improves trend alignment relative to classical ARX ($0.771 \pm 0.123$). These lead--lag patterns help reconcile the MAPE and MPE results and reinforce a consistent finding throughout the analysis: simply adding contextual signals does not automatically improve forecasting performance.

Although most forecasts lag observed changes, there are rare cases in which exogenous information eliminates this delay. As reported in Appendix~C, Hybrid ARX achieves zero-lag alignment in Berks County ($\rho^\star = 0.80$), and ARX achieves zero-lag alignment in Philadelphia County ($\rho^\star = 0.65$). Both counties belong to the high-intensity tertile. While these cases are uncommon, they highlight the best-case boundary of short-horizon forecasting where structured exogenous signals are reducing temporal delay under favorable data conditions.

\section{Discussion}

This study examined whether large language models (LLMs) can improve hospitalization forecasting compared to classical time-series methods, either through the incorporation of contextual signals or through direct prompting.

Prompt-only LLM forecasts were competitive in terms of MAPE, which measures forecast accuracy, and exhibited minimal bias, indicating that the model did not consistently overpredict or underpredict hospitalization levels. Although the LLM forecasts typically lagged observed hospitalization changes by approximately one week, they were still able to capture the overall trend of increasing and decreasing hospitalizations, in some settings with comparable or lower variance than Lag-1 models. This supports the capability of LLMs to extract temporal structure from simple numeric sequences and track the direction and magnitude of hospitalization trends, despite not being specialized time-series models \citep{jin2024llmtimeseries}. In practice, this capability is valuable for clinicians and public health officials who require forecasting models that can adapt in rapidly evolving situations, where frequent model recalibration may be costly and time-intensive. 

We also identified a possible mechanism for LLMs to be effective in traditional forecasting pipelines in healthcare. While LLMs may not yet reliably function as standalone “black-box” forecasters, they function effectively as contextual signal extractors within structured forecasting pipelines. Classical ARX models performed poorly across several evaluation metrics; however, \textsc{HybridARX} consistently outperformed classical ARX across all metrics.

One possible explanation is that LLM-generated contextual signals help smooth or regularize noisy external indicators before they are incorporated into the forecasting model. Because leading indicators anticipate shifts in hospitalization demand, improving the stability of these signals is particularly important. Many public health indicators are inconsistently reported and partially missing, which can reduce their usefulness when directly incorporated into regression-based models. Transforming noisy leading signals into more structured predictors may allow forecasting models to better leverage this information. These results suggest that LLMs are currently most valuable in this role. This is further supported by the improvements of \textsc{HybridARX} relative to classical ARX in lower-intensity settings, where signals are inherently sparse.

However, these results also highlight that simple baselines such as Lag-1 persistence and exponential smoothing should not be overlooked. While Lag-1 produced accurate forecasts with minimal bias, exponential smoothing, despite its strong accuracy, exhibited consistent negative bias, systematically underestimating hospitalization levels. This highlights an important operational consideration: a model that achieves strong accuracy may still be problematic if it systematically underestimates demand or misrepresents critical turning points. These periods are often when healthcare decision-makers require the most support in planning staffing and resource allocation. If forecasts are misleading during these critical periods, they may place additional strain on healthcare systems.

Finally, the performance of the ARX model suggests that incorporating temporal structure alone is not sufficient. The addition of contextual signals does not automatically improve forecasting accuracy, and across the diverse data sources considered, simply including more inputs was not sufficient to improve performance \citep{williams2025contextiskey}. Advancing hospitalization forecasting therefore requires moving beyond short-term trend extrapolation toward identifying contextual signals that enable earlier detection of shifts in patient surges. LLMs may play a significant role in this transition due to their ability to reduce noise in heterogeneous contextual data and transform these signals into structured predictors suitable for forecasting pipelines.

\section{Limitations}

The lead--lag analysis showed that none of the evaluated models could consistently anticipate changes in hospitalization demand. This suggests that short-horizon hospitalization forecasting primarily reacts to observed data, tracking evolving trends but failing to detect early shifts. More broadly, this reflects a fundamental challenge: hospital admissions are themselves lagging indicators of disease spread, and short-horizon models rely heavily on recent observations. Additionally, our approach does not explicitly produce probabilistic uncertainty estimates (e.g., prediction intervals such as 95\% confidence intervals), and instead captures only epistemic uncertainty through run-to-run variability in LLM-based forecasts.

In this setting, the limitation is not solely due to model design, but also the availability and timing of informative signals. Although LLMs are capable of achieving performance comparable to specialized time-series forecasters, their effectiveness depends on the quality of input signals and whether those signals meaningfully precede changes in hospitalization demand. While search-based behavioral signals consistently precede hospitalizations, other contextual signals exhibit weak or inconsistent leading behavior. This highlights a key limitation in the availability of reliable leading indicators, suggesting that the primary challenge lies not only in model design but also in identifying signals that truly anticipate hospitalization demand.

In addition, our analysis is conducted using county-level aggregated data, which provides a consistent and comparable view across regions but does not fully reflect the data-sharing realities of hospital systems. In practice, hospitals operate within fragmented data environments, where information is distributed across facilities with heterogeneous reporting standards, privacy constraints, and access limitations. This discrepancy may affect how forecasting models perform in real-world hospital deployment.

\section{Conclusion}

Reliable and timely hospitalization forecasting is crucial in aiding clinical decision-makers to allocate beds, ventilators, and manage ICU capacity. This, in turn, promotes the well-being of healthcare staff by reducing burnout, helps ensure that patient safety is not compromised, and supports sustainable allocation of hospital resources.

In this work, we introduced \textsc{HybridARX}, a context-augmented hybrid forecasting framework that utilizes LLMs to encode and smooth unstructured, diverse contextual signals within established time-series models. Our findings are situated within the broader context of healthcare systems, where forecasting is often conducted under heterogeneous capacity constraints and fragmented data environments, and demonstrate that effective hospitalization forecasting is not limited to accuracy alone, but also depends on stability, reduced bias, and strong temporal alignment.

While this work focuses on predictive performance, an important direction for future work is identifying which exogenous signals directly influence hospitalization outcomes. Understanding which signals meaningfully precede changes in demand can help narrow the factors that hospitals should prioritize, enabling more actionable and decision-aware forecasting for epidemic response.

\bibliography{sample}

\section{Appendix A: Prompt Templates}

\subsection{Prompt-Only}
The prompt-only approach predicts next-week COVID-19 hospitalizations using the previous eight weeks of hospitalization data. No exogenous indicators (e.g., ICU capacity, ventilator utilization, or search trends) are provided. The prompt wording and structure are fixed across all counties and weeks. Only geography, dates, and other data vary.

\begin{quote}
\small
\textbf{Template (placeholders shown in angle brackets):}

\medskip
Given the last 8 weekly observations for region \texttt{<Geography>}:

\medskip
\texttt{<RecentData>}

\medskip
Predict next week's value (week ending \texttt{<PredictionDate>}) for the numeric series y.

\medskip
Return exactly one line:\\
\texttt{y: <number>}
\end{quote}
\medskip

Return exactly three lines:\\
\texttt{X\_B: <number>}\\
\texttt{X\_V: <number>}\\
\texttt{s\_t: <number>}
\textit{Inserted values:} \texttt{<Geography>} specifies the county name (e.g., ``Adams'').  \\

\texttt{<PredictionDate>} corresponds to the week-ending date of the forecast period. \texttt{<RecentData>} is a chronologically ordered block containing the previous eight weeks of 14-day average COVID-19 hospitalization values, formatted as date-value pairs (e.g., ``2020-04-06: y=0.46'').

\subsection{\textsc{HybridARX} LLM Template}
\label{app:context-augmented-template}

In the \textsc{HybridARX}
pipeline, the LLM is used only to forecast selected contextual indicators for the prediction week. These LLM-derived indicators are then incorporated as exogenous variables in downstream time-series models to generate hospitalization forecasts. The LLM does not directly predict hospitalizations in this pipeline.

\begin{quote}
\small
\textbf{Template (placeholders shown in angle brackets):}

\medskip
Given the last 8 weekly observations for region \texttt{<Geography>}:

\medskip
\texttt{<RecentData>}

\medskip
Predict next week's values (week ending \texttt{<PredictionDate>}) for three numeric series: $X_B$, $X_V$, $s_t$

\medskip
Return exactly three lines:\\
\texttt{X\_B: <number>}\\
\texttt{X\_V: <number>}\\
\texttt{s\_t: <number>}
\end{quote}

\textit{Inserted values:} \texttt{<Geography>} specifies the county name. 

\texttt{<PredictionDate>} corresponds to the week-ending date of the forecast period. \texttt{<RecentData>} is a chronologically ordered block containing the previous eight weeks of contextual indicator values, formatted as date-value tuples:
\begin{itemize}
    \item \texttt{X\_B}: Adult ICU beds total (weekly mean)
    \item \texttt{X\_V}: COVID-19 patients on ventilators (weekly mean)
    \item \texttt{s\_t}: Anosmia/ageusia Google search volume (weekly sum)
\end{itemize}
\noindent The resulting predictions $(\hat{X}_{B,t+1}, \hat{X}_{V,t+1}, \hat{s}_{t+1})$ are passed to the rolling-window ARX and LinReg forecasting pipeline as exogenous inputs to produce the final hospitalization forecast $\hat{y}_{t+1}$.

%%%%%%%%%%%%%%%%%%%%%%%%%%%%%%%%%%%%%%%%%%%%%%%%%%%%%%%%%%%%%%%%%%%%%%
\section{Appendix B: Model Hyperparameters and Implementation Details}
\label{app:model-hyperparameters}
%%%%%%%%%%%%%%%%%%%%%%%%%%%%%%%%%%%%%%%%%%%%%%%%%%%%%%%%%%%%%%%%%%%%%%

This appendix summarizes the hyperparameters and implementation choices used for the classical time-series baselines, prompt-only LLM, and \textsc{HybridARX} approach. 

\subsection{Classical Time-Series Baselines}
\label{app:classical-baselines}

All classical baselines are implemented using a rolling-window framework with a fixed history length of $L=8$ weeks. Models are fit independently for each county and each forecast origin.

\paragraph{Lag-1 Baseline.}
A naïve persistence model requiring no hyperparameters. The forecast for week $t+1$ is simply the observed value at week $t$:
\[
\hat{y}_{t+1} = y_t.
\]

\paragraph{Autoregressive Model (AR(1)).}
A first-order AR model with a constant intercept.
\begin{itemize}
    \item \textbf{Lag order.} \texttt{lags=1}.
    \item \textbf{Trend.} Constant intercept (\texttt{trend='c'}).
    \item \textbf{Training window.} Previous 8 weekly observations.
\end{itemize}

\paragraph{Exponential Smoothing.}
Additive Holt's Linear Trend:
\begin{itemize}
    \item \textbf{Trend.} Additive (\texttt{trend='add'}).
    \item \textbf{Seasonality.} Disabled (\texttt{seasonal=None}) in the reported experiments due to the fixed 8-week rolling window.
    \item \textbf{Initialization.} Estimated automatically.
    \item \textbf{Optimization.} Enabled during model fitting.
\end{itemize}

\paragraph{AR Model with Exogenous Inputs (ARX).}
An AR(1) model augmented with exogenous regressors
\begin{itemize}
    \item \textbf{Lag order.} \texttt{lags=1}.
    \item \textbf{Trend.} Constant intercept (\texttt{trend='c'}).
    \item \textbf{Exogenous variables.} Three contextual indicators:
    \begin{itemize}
        \item $X_B$: Adult ICU beds total (weekly mean),
        \item $X_V$: COVID-19 patients on ventilators (weekly mean),
        \item $s_t$: Anosmia/ageusia search volume (weekly sum).
    \end{itemize}
    \item \textbf{Training window.} Previous 8 weekly observations.
    \item \textbf{Missing value handling.} Column-wise mean imputation within the training window, with any remaining missing values set to zero.
\end{itemize}

The 8-week training window provides sufficient degrees of freedom to estimate the intercept, a single AR lag, and three exogenous regressors without overfitting.

\paragraph{Post-Processing.}
All baseline forecasts are clipped to be non-negative, as negative hospitalization counts are not meaningful.

\subsection{Prompt-Only LLM}
\label{app:prompt-only-llm}

\begin{itemize}
    \item \textbf{Model.} \texttt{gpt-5-2025-08-07}, accessed via the OpenAI API.
    \item \textbf{Temperature.} Default setting.
    \item \textbf{Context window.} The previous $L=8$ weekly observations of $y_t$, where $y_t$ denotes the 14-day average number of hospitalized COVID-19 patients.
    \item \textbf{Output parsing.} The predicted value is extracted using a regular expression matching a labeled numeric output (\texttt{y: <number>}). If parsing fails, the prompt is retried once with stricter formatting instructions, or else it is returned as missing.
    \item \textbf{Runs.} Three independent runs are performed for each county--week to characterize run-to-run variability.
    \item \textbf{Post-processing.} All predictions are clipped to be non-negative.
\end{itemize}

\subsection{Context-Augmented Hybrid Model [Multivariate] (\textsc{HybridARX})}
\label{app:hybrid}

\subsubsection{LLM-Based Contextual Forecasting}

\begin{itemize}
    \item \textbf{Model.} \texttt{gpt-5-2025-08-07} accessed via the OpenAI API.
    \item \textbf{Temperature.} Default setting.
    \item \textbf{Context window.} The previous $L=8$ weekly observations of $(y_t, X_{B,t}, X_{V,t}, s_t)$.
    \item \textbf{Predicted indicators.}
    \begin{itemize}
        \item $\hat{X}_{B,t+1}$: Adult ICU beds total,
        \item $\hat{X}_{V,t+1}$: COVID-19 patients on ventilators,
        \item $\hat{s}_{t+1}$: Anosmia/ageusia search volume.
    \end{itemize}
    \item \textbf{Output parsing.} Numeric values are extracted using regular expressions matching labeled outputs (\texttt{X\_B:}, \texttt{X\_V:}, \texttt{s\_t:}). If any label is missing, the prediction is treated as missing.
    \item \textbf{Retry logic.} When parsing fails, the prompt is resubmitted once with stricter formatting instructions.
    \item \textbf{Runs.} Three independent runs are performed to characterize run-to-run variability.
\end{itemize}

\subsubsection{Statistical Forecasting with LLM-Predicted Context}

\paragraph{ARX with LLM-Derived Context.}
\begin{itemize}
    \item \textbf{Model.} First-order AR model with exogenous regressors (ARX).
    \item \textbf{Training data.} The previous $L$ weeks of hospitalizations $y_{t-L:t}$ and contextual indicators $X_{t-L:t}$ from the county--weekly panel.
    \item \textbf{Prediction input.} LLM-predicted ICU beds, ventilator utilization, and anosmia search volume $(\hat{X}_{B,t+1}, \hat{X}_{V,t+1}, \hat{s}_{t+1})$.
    \item \textbf{Missing values.} Column-wise mean imputation within the training window. Remaining missing values are set to zero.
\end{itemize}

\paragraph{Linear Regression with LLM-Derived Context.}
\begin{itemize}
    \item \textbf{Model.} Ordinary least squares linear regression with intercept.
    \item \textbf{Training data.} Historical hospitalizations $y_{t-L:t}$ regressed on contextual indicators $X_{t-L:t}$.
    \item \textbf{Prediction input.} LLM-predicted contextual indicators $(\hat{X}_{B,t+1}, \hat{X}_{V,t+1}, \hat{s}_{t+1})$.
\end{itemize}

\paragraph{Post-Processing.}
All Stage~2 forecasts are clipped to non-negative values.

\clearpage %

\section{Appendix C: Additional Tables}
\label{app:tables}

% ---------------- LOW ----------------
\begin{table*}[p]
\centering
\caption{MAPE: low-intensity counties (per-county mean percent error $\pm$ SD).}
\label{tab:mape-low-intensity}
{\footnotesize
\setlength{\tabcolsep}{3pt}
\renewcommand{\arraystretch}{0.95}
\resizebox{\textwidth}{!}{%
\begin{tabular}{lrrrrrrr}
\toprule
County & Lag-1 & AR(1) & ES & ARX & LLM & Hybrid ARX & Hybrid LR \\
\midrule
Armstrong County & $25.5 \pm 22.9$ & $32.9 \pm 32.8$ & $29.3 \pm 27.9$ & $36.1 \pm 35.7$ & $25.4 \pm 23.7$ & $31.1 \pm 32.1$ & $46.5 \pm 46.8$ \\
Bedford County & $28.7 \pm 31.0$ & $48.4 \pm 140.9$ & $29.5 \pm 28.2$ & $48.8 \pm 139.1$ & $30.1 \pm 31.5$ & $46.6 \pm 139.4$ & $40.0 \pm 43.0$ \\
Carbon County & $33.3 \pm 40.9$ & $36.7 \pm 42.8$ & $34.3 \pm 40.3$ & $42.4 \pm 39.8$ & $36.4 \pm 41.0$ & $35.6 \pm 31.9$ & $59.8 \pm 59.5$ \\
Clarion County & $25.9 \pm 32.0$ & $34.4 \pm 40.1$ & $35.9 \pm 50.0$ & $37.2 \pm 46.9$ & $29.0 \pm 37.0$ & $34.0 \pm 43.9$ & $42.1 \pm 48.1$ \\
Clinton County & $22.4 \pm 38.1$ & $54.2 \pm 174.4$ & $26.1 \pm 41.7$ & $56.8 \pm 165.5$ & $24.4 \pm 39.2$ & $51.6 \pm 165.0$ & $45.0 \pm 47.3$ \\
Columbia County & $25.6 \pm 27.0$ & $30.2 \pm 31.4$ & $27.7 \pm 28.4$ & $29.2 \pm 36.8$ & $27.8 \pm 36.8$ & $29.6 \pm 37.5$ & $39.6 \pm 50.3$ \\
Elk County & $19.9 \pm 23.3$ & $25.3 \pm 32.6$ & $22.1 \pm 24.8$ & $23.0 \pm 23.2$ & $21.7 \pm 23.8$ & $22.7 \pm 23.8$ & $26.0 \pm 25.1$ \\
Fulton County & $23.3 \pm 21.9$ & $34.9 \pm 35.0$ & $26.7 \pm 29.4$ & $39.8 \pm 42.6$ & $24.9 \pm 26.8$ & $31.1 \pm 32.7$ & $40.3 \pm 37.4$ \\
Greene County & $21.4 \pm 22.0$ & $26.0 \pm 28.2$ & $24.0 \pm 24.8$ & $28.6 \pm 35.2$ & $24.1 \pm 25.2$ & $27.8 \pm 31.9$ & $32.5 \pm 31.5$ \\
Huntingdon County & $18.3 \pm 22.9$ & $25.2 \pm 34.1$ & $23.1 \pm 23.1$ & $28.9 \pm 38.3$ & $21.8 \pm 26.4$ & $25.0 \pm 35.8$ & $33.5 \pm 36.6$ \\
Jefferson County & $23.1 \pm 25.4$ & $29.3 \pm 37.2$ & $25.3 \pm 25.1$ & $36.3 \pm 45.6$ & $23.2 \pm 23.3$ & $27.3 \pm 34.1$ & $41.5 \pm 47.8$ \\
Lawrence County & $25.0 \pm 23.5$ & $28.7 \pm 27.1$ & $28.7 \pm 29.8$ & $36.5 \pm 44.5$ & $25.2 \pm 22.5$ & $31.4 \pm 51.1$ & $31.5 \pm 32.1$ \\
McKean County & $21.4 \pm 24.7$ & $30.2 \pm 44.2$ & $21.5 \pm 27.3$ & $34.8 \pm 56.7$ & $20.1 \pm 26.6$ & $26.0 \pm 42.5$ & $36.0 \pm 43.8$ \\
Northumberland County & $21.5 \pm 18.3$ & $24.8 \pm 21.6$ & $22.8 \pm 21.5$ & $28.4 \pm 23.7$ & $23.7 \pm 22.5$ & $24.3 \pm 23.4$ & $39.3 \pm 41.5$ \\
Potter County & $24.3 \pm 25.7$ & $32.4 \pm 37.5$ & $24.6 \pm 23.8$ & $35.0 \pm 41.0$ & $26.4 \pm 25.9$ & $31.9 \pm 36.7$ & $43.9 \pm 56.7$ \\
Susquehanna County & $27.9 \pm 27.1$ & $33.9 \pm 34.9$ & $35.7 \pm 34.6$ & $34.9 \pm 33.6$ & $31.1 \pm 32.5$ & $33.9 \pm 33.8$ & $36.3 \pm 36.3$ \\
Tioga County & $25.4 \pm 29.4$ & $34.7 \pm 36.8$ & $29.7 \pm 31.1$ & $35.6 \pm 32.2$ & $26.6 \pm 27.1$ & $27.8 \pm 30.2$ & $35.6 \pm 39.8$ \\
Warren County & $24.6 \pm 29.2$ & $35.4 \pm 43.8$ & $22.3 \pm 25.9$ & $38.3 \pm 46.4$ & $22.7 \pm 26.2$ & $31.9 \pm 38.1$ & $64.8 \pm 91.7$ \\
Wayne County & $24.0 \pm 22.2$ & $28.4 \pm 27.2$ & $22.5 \pm 21.4$ & $38.2 \pm 63.4$ & $26.3 \pm 27.3$ & $30.8 \pm 29.5$ & $41.8 \pm 47.5$ \\
Wyoming County & $14.4 \pm 21.9$ & $22.9 \pm 51.6$ & $17.4 \pm 23.7$ & $26.0 \pm 41.5$ & $17.8 \pm 24.1$ & $22.9 \pm 39.5$ & $22.3 \pm 23.7$ \\
\bottomrule
\end{tabular}
}
}
\end{table*}
% ---------------- MID ----------------
\begin{table*}[p]
\centering
\caption{MAPE: mid-intensity counties (per-county mean percent error $\pm$ SD).}
\label{tab:mape-mid-intensity}
{\footnotesize
\setlength{\tabcolsep}{3pt}
\renewcommand{\arraystretch}{0.95}
\resizebox{\textwidth}{!}{%
\begin{tabular}{lrrrrrrr}
\toprule
County & Lag-1 & AR(1) & ES & ARX & LLM & Hybrid ARX & Hybrid LR \\
\midrule
Adams County & $18.4 \pm 18.0$ & $22.6 \pm 20.8$ & $19.4 \pm 20.0$ & $32.0 \pm 32.8$ & $20.1 \pm 20.8$ & $24.2 \pm 23.0$ & $21.1 \pm 21.1$ \\
Beaver County & $22.3 \pm 21.5$ & $28.0 \pm 30.2$ & $24.0 \pm 26.3$ & $35.7 \pm 44.5$ & $29.0 \pm 30.3$ & $26.2 \pm 25.5$ & $31.1 \pm 33.1$ \\
Bradford County & $23.1 \pm 31.1$ & $24.5 \pm 28.4$ & $19.9 \pm 21.6$ & $23.4 \pm 29.0$ & $21.3 \pm 25.8$ & $22.5 \pm 27.6$ & $38.9 \pm 86.9$ \\
Butler County & $24.7 \pm 31.1$ & $28.7 \pm 30.5$ & $27.6 \pm 41.2$ & $33.6 \pm 42.5$ & $25.5 \pm 35.9$ & $26.3 \pm 33.0$ & $31.6 \pm 40.7$ \\
Cambria County & $16.6 \pm 12.4$ & $17.7 \pm 14.4$ & $15.4 \pm 15.8$ & $20.3 \pm 20.6$ & $16.1 \pm 12.2$ & $16.2 \pm 13.3$ & $22.2 \pm 21.2$ \\
Centre County & $18.7 \pm 19.9$ & $21.4 \pm 21.4$ & $15.8 \pm 16.4$ & $25.0 \pm 31.5$ & $18.8 \pm 23.6$ & $19.8 \pm 26.3$ & $35.7 \pm 50.0$ \\
Clearfield County & $19.7 \pm 19.5$ & $21.3 \pm 20.1$ & $20.9 \pm 23.4$ & $27.5 \pm 32.8$ & $20.3 \pm 19.5$ & $19.9 \pm 19.4$ & $26.6 \pm 22.8$ \\
Crawford County & $21.6 \pm 23.6$ & $26.8 \pm 35.1$ & $22.3 \pm 33.8$ & $27.0 \pm 34.0$ & $23.2 \pm 36.7$ & $21.7 \pm 30.8$ & $28.9 \pm 37.4$ \\
Fayette County & $19.8 \pm 16.5$ & $23.6 \pm 21.6$ & $18.3 \pm 20.6$ & $27.5 \pm 31.6$ & $17.1 \pm 17.1$ & $21.6 \pm 23.5$ & $35.6 \pm 37.7$ \\
Indiana County & $24.6 \pm 23.2$ & $33.2 \pm 29.3$ & $26.6 \pm 24.4$ & $33.8 \pm 30.4$ & $23.8 \pm 22.0$ & $24.8 \pm 20.0$ & $39.1 \pm 49.4$ \\
Lebanon County & $18.9 \pm 18.1$ & $20.2 \pm 19.8$ & $17.0 \pm 19.0$ & $20.7 \pm 28.1$ & $18.6 \pm 16.1$ & $19.0 \pm 23.4$ & $30.3 \pm 46.7$ \\
Lycoming County & $24.8 \pm 27.1$ & $26.8 \pm 26.3$ & $24.6 \pm 25.6$ & $32.5 \pm 38.0$ & $23.6 \pm 23.0$ & $25.8 \pm 25.9$ & $44.6 \pm 73.8$ \\
Mercer County & $26.0 \pm 23.0$ & $31.1 \pm 30.5$ & $23.5 \pm 24.5$ & $34.9 \pm 33.6$ & $23.6 \pm 21.4$ & $29.6 \pm 30.6$ & $52.3 \pm 89.5$ \\
Mifflin County & $21.9 \pm 18.5$ & $24.7 \pm 24.1$ & $24.7 \pm 22.2$ & $30.1 \pm 32.9$ & $23.0 \pm 20.6$ & $26.1 \pm 29.5$ & $37.1 \pm 43.2$ \\
Monroe County & $24.1 \pm 28.5$ & $30.5 \pm 31.3$ & $22.8 \pm 31.6$ & $40.0 \pm 63.0$ & $22.3 \pm 23.9$ & $30.7 \pm 46.3$ & $41.8 \pm 61.3$ \\
Northampton County & $23.9 \pm 23.3$ & $30.0 \pm 25.6$ & $23.4 \pm 20.0$ & $30.0 \pm 27.9$ & $20.2 \pm 17.2$ & $23.4 \pm 22.4$ & $53.4 \pm 71.6$ \\
Schuylkill County & $19.8 \pm 15.6$ & $20.1 \pm 16.8$ & $18.3 \pm 18.8$ & $22.8 \pm 23.9$ & $18.9 \pm 17.6$ & $18.4 \pm 17.9$ & $38.7 \pm 53.0$ \\
Somerset County & $28.9 \pm 45.9$ & $37.6 \pm 40.4$ & $27.9 \pm 38.9$ & $37.4 \pm 44.1$ & $29.3 \pm 44.6$ & $31.3 \pm 36.5$ & $57.7 \pm 80.2$ \\
Union County & $26.1 \pm 21.5$ & $30.5 \pm 24.8$ & $26.2 \pm 25.2$ & $44.1 \pm 67.2$ & $24.9 \pm 23.4$ & $32.5 \pm 30.1$ & $44.8 \pm 48.5$ \\
Venango County & $29.1 \pm 26.0$ & $39.4 \pm 38.1$ & $32.3 \pm 26.4$ & $44.9 \pm 58.5$ & $32.2 \pm 31.8$ & $38.5 \pm 48.6$ & $51.5 \pm 81.9$ \\
\bottomrule
\end{tabular}
}
}
\end{table*}
% ---------------- HIGH ----------------
\begin{table*}[p]
\centering
\caption{MAPE: high-intensity counties (per-county mean percent error $\pm$ SD).}
\label{tab:mape-high-intensity}
{\footnotesize
\setlength{\tabcolsep}{3pt}
\renewcommand{\arraystretch}{0.95}
\resizebox{\textwidth}{!}{%
\begin{tabular}{lrrrrrrr}
\toprule
County & Lag-1 & AR(1) & ES & ARX & LLM & Hybrid ARX & Hybrid LR \\
\midrule
\mbox{Allegheny County} & $17.1 \pm 10.4$ & $18.8 \pm 12.4$ & $12.5 \pm 15.2$ & $15.8 \pm 19.5$ & $19.6 \pm 13.0$ & $15.6 \pm 15.0$ & $19.8 \pm 17.2$ \\
\mbox{Berks County} & $18.4 \pm 16.1$ & $20.5 \pm 22.2$ & $14.2 \pm 19.7$ & $28.2 \pm 58.5$ & $16.6 \pm 17.9$ & $17.7 \pm 20.1$ & $29.6 \pm 44.6$ \\
\mbox{Blair County} & $23.4 \pm 23.6$ & $26.6 \pm 20.6$ & $22.3 \pm 25.1$ & $36.7 \pm 43.2$ & $23.2 \pm 22.0$ & $27.1 \pm 28.6$ & $39.2 \pm 77.1$ \\
\mbox{Bucks County} & $21.1 \pm 21.6$ & $23.2 \pm 30.6$ & $18.6 \pm 22.7$ & $24.1 \pm 36.2$ & $20.3 \pm 18.6$ & $19.8 \pm 22.5$ & $35.8 \pm 44.9$ \\
\mbox{Chester County} & $17.7 \pm 12.1$ & $18.1 \pm 15.7$ & $14.1 \pm 12.5$ & $20.7 \pm 19.7$ & $15.5 \pm 14.0$ & $16.9 \pm 15.8$ & $26.1 \pm 26.8$ \\
\mbox{Cumberland County} & $16.1 \pm 12.1$ & $16.0 \pm 11.9$ & $15.1 \pm 15.5$ & $24.0 \pm 25.4$ & $17.9 \pm 13.4$ & $18.7 \pm 14.1$ & $20.9 \pm 20.4$ \\
\mbox{Dauphin County} & $13.0 \pm 12.5$ & $15.4 \pm 15.7$ & $12.0 \pm 15.5$ & $21.1 \pm 27.5$ & $15.3 \pm 15.3$ & $13.9 \pm 17.6$ & $17.2 \pm 19.4$ \\
\mbox{Delaware County} & $17.8 \pm 13.8$ & $22.4 \pm 19.0$ & $13.7 \pm 16.1$ & $22.7 \pm 24.1$ & $20.7 \pm 16.0$ & $19.8 \pm 18.3$ & $26.0 \pm 22.8$ \\
\mbox{Erie County} & $17.9 \pm 19.2$ & $23.4 \pm 23.5$ & $13.7 \pm 16.8$ & $23.8 \pm 24.0$ & $23.9 \pm 29.1$ & $21.6 \pm 23.4$ & $30.5 \pm 40.9$ \\
\mbox{Franklin County} & $19.6 \pm 15.6$ & $22.5 \pm 20.1$ & $17.9 \pm 21.7$ & $23.0 \pm 25.9$ & $21.8 \pm 18.4$ & $19.7 \pm 21.3$ & $26.3 \pm 25.0$ \\
\mbox{Lackawanna County} & $16.6 \pm 12.3$ & $18.8 \pm 16.5$ & $15.2 \pm 13.3$ & $27.4 \pm 35.2$ & $21.8 \pm 17.1$ & $20.7 \pm 19.1$ & $22.3 \pm 16.2$ \\
\mbox{Lancaster County} & $16.2 \pm 12.2$ & $18.5 \pm 15.6$ & $14.5 \pm 17.1$ & $19.6 \pm 20.5$ & $19.6 \pm 14.4$ & $16.6 \pm 16.8$ & $20.3 \pm 15.9$ \\
\mbox{Lehigh County} & $14.9 \pm 9.1$ & $13.6 \pm 10.4$ & $8.6 \pm 8.9$ & $14.2 \pm 19.2$ & $16.4 \pm 12.7$ & $13.4 \pm 13.4$ & $19.1 \pm 12.6$ \\
\mbox{Luzerne County} & $20.9 \pm 19.7$ & $24.8 \pm 24.8$ & $22.1 \pm 26.9$ & $26.4 \pm 30.7$ & $25.0 \pm 26.0$ & $23.4 \pm 23.1$ & $30.4 \pm 34.4$ \\
\mbox{Montgomery County} & $16.9 \pm 13.1$ & $17.2 \pm 15.1$ & $12.5 \pm 15.2$ & $14.5 \pm 15.4$ & $16.3 \pm 12.3$ & $15.3 \pm 15.8$ & $20.4 \pm 16.4$ \\
\mbox{Montour County} & $17.9 \pm 22.6$ & $23.5 \pm 38.6$ & $19.4 \pm 30.5$ & $30.1 \pm 51.2$ & $19.6 \pm 26.4$ & $23.0 \pm 40.3$ & $35.0 \pm 41.1$ \\
\mbox{Philadelphia County} & $14.3 \pm 9.2$ & $13.5 \pm 10.1$ & $8.0 \pm 6.2$ & $12.4 \pm 18.1$ & $14.8 \pm 12.0$ & $12.6 \pm 11.0$ & $15.5 \pm 12.0$ \\
\mbox{Washington County} & $20.9 \pm 18.1$ & $24.0 \pm 21.5$ & $19.5 \pm 17.7$ & $29.7 \pm 36.0$ & $25.0 \pm 17.9$ & $24.8 \pm 21.1$ & $29.5 \pm 31.0$ \\
\mbox{Westmoreland County} & $21.3 \pm 22.4$ & $23.7 \pm 29.3$ & $22.4 \pm 33.3$ & $84.1 \pm 582.8$ & $21.7 \pm 30.0$ & $17.6 \pm 19.0$ & $31.3 \pm 43.5$ \\
\mbox{York County} & $15.1 \pm 13.3$ & $17.8 \pm 24.0$ & $13.6 \pm 18.5$ & $22.9 \pm 38.7$ & $16.5 \pm 19.1$ & $18.7 \pm 21.9$ & $20.1 \pm 20.5$ \\
\bottomrule
\end{tabular}%
}
}
\end{table*}

\begin{table*}[p]
\centering
\caption{MPE: low-intensity counties (per-county mean percent error $\pm$ SD).}
\label{tab:mpe-low-intensity}
\scriptsize
\setlength{\tabcolsep}{3pt}
\resizebox{\textwidth}{!}{%
\begin{tabular}{lrrrrrrr}
\toprule
County & Lag-1 & AR(1) & ES & ARX & LLM & Hybrid ARX & Hybrid LR \\
\midrule
\mbox{Armstrong County} & $+2.1 \pm 34.3$ & $+5.8 \pm 46.3$ & $-3.7 \pm 40.4$ & $+7.6 \pm 50.3$ & $+0.8 \pm 34.8$ & $+6.1 \pm 44.4$ & $+13.6 \pm 64.7$ \\
\mbox{Bedford County} & $+0.6 \pm 42.3$ & $+20.9 \pm 147.6$ & $+3.6 \pm 40.8$ & $+21.0 \pm 145.9$ & $+5.0 \pm 43.4$ & $+19.7 \pm 145.7$ & $+2.4 \pm 58.8$ \\
\mbox{Carbon County} & $+4.1 \pm 52.7$ & $+9.4 \pm 55.8$ & $-0.1 \pm 53.0$ & $+11.5 \pm 57.1$ & $+7.7 \pm 54.4$ & $+3.7 \pm 47.8$ & $+17.9 \pm 82.7$ \\
\mbox{Clarion County} & $+3.4 \pm 41.2$ & $+9.5 \pm 52.0$ & $+5.8 \pm 61.4$ & $+7.6 \pm 59.5$ & $+8.0 \pm 46.5$ & $+6.8 \pm 55.3$ & $+15.6 \pm 62.1$ \\
\mbox{Clinton County} & $+2.0 \pm 44.2$ & $+31.2 \pm 180.0$ & $+2.1 \pm 49.2$ & $+29.5 \pm 172.5$ & $+4.0 \pm 46.1$ & $+28.1 \pm 170.7$ & $+9.3 \pm 64.8$ \\
\mbox{Columbia County} & $-0.3 \pm 37.3$ & $+1.4 \pm 43.6$ & $-2.0 \pm 39.8$ & $+5.2 \pm 46.8$ & $+2.5 \pm 46.2$ & $+2.5 \pm 47.8$ & $+5.5 \pm 63.9$ \\
\mbox{Elk County} & $+0.1 \pm 30.7$ & $+7.7 \pm 40.6$ & $-3.4 \pm 33.1$ & $-3.0 \pm 32.6$ & $+0.9 \pm 32.2$ & $-4.3 \pm 32.7$ & $-3.1 \pm 36.1$ \\
\mbox{Fulton County} & $+2.4 \pm 32.0$ & $+13.2 \pm 47.7$ & $-1.4 \pm 39.7$ & $+16.2 \pm 56.1$ & $+2.9 \pm 36.6$ & $+10.3 \pm 44.1$ & $+10.9 \pm 54.0$ \\
\mbox{Greene County} & $+0.6 \pm 30.7$ & $+9.6 \pm 37.2$ & $+0.9 \pm 34.6$ & $+13.9 \pm 43.2$ & $+3.7 \pm 34.8$ & $+10.7 \pm 41.0$ & $+5.0 \pm 45.1$ \\
\mbox{Huntingdon County} & $-0.2 \pm 29.4$ & $+7.9 \pm 41.7$ & $-1.2 \pm 32.8$ & $+7.6 \pm 47.5$ & $+1.3 \pm 34.3$ & $+1.9 \pm 43.7$ & $+3.3 \pm 49.6$ \\
\mbox{Jefferson County} & $-0.5 \pm 34.4$ & $+6.1 \pm 47.0$ & $-5.1 \pm 35.4$ & $+13.5 \pm 56.8$ & $+0.2 \pm 33.0$ & $+6.3 \pm 43.3$ & $+10.2 \pm 62.7$ \\
\mbox{Lawrence County} & $+0.8 \pm 34.4$ & $+6.5 \pm 39.0$ & $+2.4 \pm 41.4$ & $-0.1 \pm 57.8$ & $-0.1 \pm 33.9$ & $-0.6 \pm 60.1$ & $+0.2 \pm 45.1$ \\
\mbox{McKean County} & $+1.1 \pm 32.7$ & $+9.7 \pm 52.7$ & $-0.6 \pm 34.8$ & $+11.5 \pm 65.6$ & $+1.8 \pm 33.4$ & $+6.3 \pm 49.5$ & $+9.1 \pm 56.0$ \\
\mbox{Northumberland County} & $+0.2 \pm 28.3$ & $+6.9 \pm 32.3$ & $+0.5 \pm 31.5$ & $+6.4 \pm 36.5$ & $+3.1 \pm 32.7$ & $+4.5 \pm 33.5$ & $+8.3 \pm 56.7$ \\
\mbox{Potter County} & $+0.8 \pm 35.4$ & $+10.6 \pm 48.6$ & $-3.0 \pm 34.2$ & $+6.4 \pm 53.7$ & $+0.4 \pm 37.1$ & $+4.0 \pm 48.6$ & $+14.9 \pm 70.3$ \\
\mbox{Susquehanna County} & $+0.4 \pm 39.0$ & $+11.8 \pm 47.4$ & $+6.8 \pm 49.3$ & $-4.3 \pm 48.4$ & $+1.9 \pm 45.1$ & $-1.6 \pm 48.0$ & $-1.4 \pm 51.5$ \\
\mbox{Tioga County} & $+2.1 \pm 38.9$ & $+6.4 \pm 50.3$ & $-6.0 \pm 42.7$ & $-2.6 \pm 48.1$ & $+0.4 \pm 38.0$ & $-2.7 \pm 41.1$ & $-2.0 \pm 53.5$ \\
\mbox{Warren County} & $+2.9 \pm 38.2$ & $+6.8 \pm 56.0$ & $-6.0 \pm 33.8$ & $+8.1 \pm 59.7$ & $+2.4 \pm 34.6$ & $-0.4 \pm 49.8$ & $+35.4 \pm 106.7$ \\
\mbox{Wayne County} & $-1.0 \pm 32.8$ & $+2.9 \pm 39.3$ & $-3.5 \pm 31.0$ & $+7.7 \pm 73.7$ & $+2.0 \pm 37.9$ & $-0.1 \pm 42.8$ & $+6.0 \pm 63.1$ \\
\mbox{Wyoming County} & $+0.3 \pm 26.3$ & $+5.6 \pm 56.2$ & $+2.5 \pm 29.4$ & $+7.4 \pm 48.5$ & $+1.2 \pm 30.0$ & $+3.2 \pm 45.6$ & $+5.0 \pm 32.2$ \\
\bottomrule
\end{tabular}%
}
\end{table*}

\begin{table*}[p]
\centering
\caption{MPE: mid-intensity counties (per-county mean percent error $\pm$ SD).}
\label{tab:mpe-mid-intensity}
\scriptsize
\setlength{\tabcolsep}{3pt}
\resizebox{\textwidth}{!}{%
\begin{tabular}{lrrrrrrr}
\toprule
County & Lag-1 & AR(1) & ES & ARX & LLM & Hybrid ARX & Hybrid LR \\
\midrule
\mbox{Adams County} & $-2.0 \pm 25.8$ & $-2.5 \pm 30.7$ & $-6.2 \pm 27.2$ & $-3.1 \pm 45.8$ & $-4.5 \pm 28.6$ & $-8.1 \pm 32.5$ & $-4.8 \pm 29.6$ \\
\mbox{Beaver County} & $+0.6 \pm 31.1$ & $+5.0 \pm 41.0$ & $-3.8 \pm 35.4$ & $+4.0 \pm 57.1$ & $+5.0 \pm 41.8$ & $-1.6 \pm 36.7$ & $+4.7 \pm 45.4$ \\
\mbox{Bradford County} & $+1.6 \pm 38.8$ & $-6.8 \pm 37.0$ & $-4.2 \pm 29.1$ & $-1.6 \pm 37.3$ & $+3.2 \pm 33.4$ & $-4.6 \pm 35.4$ & $+14.6 \pm 94.2$ \\
\mbox{Butler County} & $+1.9 \pm 39.8$ & $+2.8 \pm 42.0$ & $+0.5 \pm 49.7$ & $+6.6 \pm 53.9$ & $+3.3 \pm 44.1$ & $-1.2 \pm 42.3$ & $+9.0 \pm 50.8$ \\
\mbox{Cambria County} & $-1.5 \pm 20.7$ & $+0.5 \pm 22.9$ & $-3.0 \pm 21.9$ & $+0.5 \pm 29.0$ & $-3.2 \pm 20.0$ & $-0.7 \pm 21.1$ & $-3.8 \pm 30.5$ \\
\mbox{Centre County} & $+0.2 \pm 27.4$ & $+0.6 \pm 30.3$ & $-3.6 \pm 22.5$ & $+2.1 \pm 40.2$ & $+2.1 \pm 30.2$ & $-0.2 \pm 33.0$ & $+11.0 \pm 60.6$ \\
\mbox{Clearfield County} & $-2.0 \pm 27.8$ & $+1.0 \pm 29.3$ & $-4.3 \pm 31.2$ & $+2.3 \pm 42.9$ & $-3.6 \pm 28.0$ & $-3.3 \pm 27.7$ & $+0.2 \pm 35.2$ \\
\mbox{Crawford County} & $-2.5 \pm 32.0$ & $+1.6 \pm 44.2$ & $-0.5 \pm 40.6$ & $+5.5 \pm 43.2$ & $+3.2 \pm 43.3$ & $+2.2 \pm 37.7$ & $+2.6 \pm 47.3$ \\
\mbox{Fayette County} & $-2.2 \pm 25.7$ & $+1.6 \pm 32.1$ & $-4.3 \pm 27.3$ & $+8.6 \pm 41.1$ & $-1.2 \pm 24.2$ & $+3.2 \pm 31.8$ & $+6.4 \pm 51.6$ \\
\mbox{Indiana County} & $+1.1 \pm 34.0$ & $+7.4 \pm 43.8$ & $-7.5 \pm 35.4$ & $+10.0 \pm 44.5$ & $+1.7 \pm 32.4$ & $+3.0 \pm 31.9$ & $+11.8 \pm 62.0$ \\
\mbox{Lebanon County} & $+0.5 \pm 26.3$ & $-3.0 \pm 28.2$ & $-5.7 \pm 24.9$ & $-2.1 \pm 34.9$ & $-1.0 \pm 24.7$ & $-0.8 \pm 30.2$ & $+10.5 \pm 54.8$ \\
\mbox{Lycoming County} & $+3.0 \pm 36.8$ & $+2.9 \pm 37.5$ & $-7.1 \pm 34.9$ & $+6.2 \pm 49.7$ & $+2.3 \pm 33.0$ & $+0.3 \pm 36.7$ & $+16.5 \pm 84.7$ \\
\mbox{Mercer County} & $+0.6 \pm 34.8$ & $+2.2 \pm 43.6$ & $-8.4 \pm 33.0$ & $-1.8 \pm 48.6$ & $+1.3 \pm 31.9$ & $-6.5 \pm 42.2$ & $+23.3 \pm 101.1$ \\
\mbox{Mifflin County} & $-2.1 \pm 28.7$ & $-0.9 \pm 34.7$ & $-2.4 \pm 33.3$ & $-2.1 \pm 44.6$ & $+1.9 \pm 30.9$ & $-2.1 \pm 39.4$ & $+6.4 \pm 56.7$ \\
\mbox{Monroe County} & $+1.5 \pm 37.4$ & $-5.8 \pm 43.5$ & $-5.8 \pm 38.6$ & $+18.2 \pm 72.5$ & $+4.0 \pm 32.5$ & $+8.3 \pm 55.0$ & $+12.3 \pm 73.3$ \\
\mbox{Northampton County} & $+3.2 \pm 33.3$ & $-5.4 \pm 39.2$ & $-9.1 \pm 29.5$ & $-3.1 \pm 41.0$ & $+1.1 \pm 26.6$ & $-6.6 \pm 31.9$ & $+27.5 \pm 85.1$ \\
\mbox{Schuylkill County} & $+0.1 \pm 25.3$ & $+2.0 \pm 26.2$ & $-5.8 \pm 25.7$ & $-0.0 \pm 33.2$ & $-2.5 \pm 25.8$ & $-2.4 \pm 25.7$ & $+12.6 \pm 64.5$ \\
\mbox{Somerset County} & $+3.9 \pm 54.2$ & $+6.3 \pm 55.0$ & $+0.3 \pm 48.0$ & $+13.4 \pm 56.3$ & $+6.9 \pm 53.0$ & $+5.3 \pm 47.9$ & $+29.1 \pm 94.6$ \\
\mbox{Union County} & $+1.4 \pm 33.9$ & $-0.0 \pm 39.5$ & $-4.7 \pm 36.1$ & $+8.6 \pm 80.1$ & $-1.3 \pm 34.2$ & $+1.8 \pm 44.4$ & $+5.8 \pm 65.9$ \\
\mbox{Venango County} & $+1.0 \pm 39.1$ & $+7.7 \pm 54.4$ & $-2.5 \pm 41.8$ & $+0.5 \pm 73.9$ & $+1.3 \pm 45.4$ & $+5.7 \pm 61.9$ & $+18.8 \pm 95.1$ \\
\bottomrule
\end{tabular}%
}
\end{table*}

\begin{table*}[p]
\centering
\caption{MPE: high-intensity counties (per-county mean percent error $\pm$ SD).}
\label{tab:mpe-high-intensity}
\scriptsize
\setlength{\tabcolsep}{3pt}
\resizebox{\textwidth}{!}{%
\begin{tabular}{lrrrrrrr}
\toprule
County & Lag-1 & AR(1) & ES & ARX & LLM & Hybrid ARX & Hybrid LR \\
\midrule
\mbox{Allegheny County} & $-0.8 \pm 20.1$ & $-0.5 \pm 22.6$ & $-6.2 \pm 18.7$ & $-0.6 \pm 25.2$ & $-5.7 \pm 22.9$ & $-3.0 \pm 21.5$ & $-3.9 \pm 26.0$ \\
\mbox{Berks County} & $+1.4 \pm 24.5$ & $-4.6 \pm 30.0$ & $-5.7 \pm 23.7$ & $-0.2 \pm 65.0$ & $-3.3 \pm 24.2$ & $-4.0 \pm 26.6$ & $+7.9 \pm 53.0$ \\
\mbox{Blair County} & $+3.1 \pm 33.2$ & $+4.0 \pm 33.5$ & $-5.9 \pm 33.1$ & $+3.4 \pm 56.8$ & $+1.7 \pm 32.0$ & $-0.8 \pm 39.5$ & $+14.4 \pm 85.4$ \\
\mbox{Bucks County} & $+2.3 \pm 30.1$ & $+3.5 \pm 38.4$ & $-9.3 \pm 27.9$ & $+4.3 \pm 43.4$ & $+1.0 \pm 27.6$ & $-2.0 \pm 30.0$ & $+9.1 \pm 56.8$ \\
\mbox{Chester County} & $+1.4 \pm 21.5$ & $-1.2 \pm 24.0$ & $-3.6 \pm 18.6$ & $-1.0 \pm 28.6$ & $-1.2 \pm 20.9$ & $-4.8 \pm 22.7$ & $+2.8 \pm 37.4$ \\
\mbox{Cumberland County} & $-0.1 \pm 20.2$ & $+0.3 \pm 20.0$ & $-6.6 \pm 20.6$ & $+0.0 \pm 35.1$ & $-0.3 \pm 22.4$ & $+0.2 \pm 23.5$ & $+3.7 \pm 29.0$ \\
\mbox{Dauphin County} & $-0.6 \pm 18.1$ & $-3.2 \pm 21.8$ & $-2.7 \pm 19.5$ & $-6.3 \pm 34.2$ & $-2.9 \pm 21.5$ & $-3.4 \pm 22.2$ & $+1.1 \pm 26.0$ \\
\mbox{Delaware County} & $+1.6 \pm 22.6$ & $-2.8 \pm 29.3$ & $-6.9 \pm 20.0$ & $-0.9 \pm 33.2$ & $-0.5 \pm 26.3$ & $-6.0 \pm 26.4$ & $+2.0 \pm 34.6$ \\
\mbox{Erie County} & $-0.5 \pm 26.4$ & $+0.3 \pm 33.2$ & $-4.3 \pm 21.2$ & $+6.6 \pm 33.2$ & $-0.2 \pm 37.8$ & $+0.8 \pm 32.0$ & $+7.3 \pm 50.7$ \\
\mbox{Franklin County} & $+1.3 \pm 25.0$ & $+1.7 \pm 30.2$ & $-6.9 \pm 27.3$ & $+0.1 \pm 34.7$ & $-0.3 \pm 28.6$ & $-1.5 \pm 29.1$ & $+2.1 \pm 36.3$ \\
\mbox{Lackawanna County} & $+0.4 \pm 20.8$ & $-1.6 \pm 25.1$ & $-3.3 \pm 20.0$ & $+2.5 \pm 44.7$ & $-4.5 \pm 27.5$ & $-1.7 \pm 28.2$ & $+1.1 \pm 27.6$ \\
\mbox{Lancaster County} & $+0.4 \pm 20.4$ & $-2.7 \pm 24.1$ & $-5.4 \pm 21.8$ & $-0.9 \pm 28.4$ & $-1.3 \pm 24.4$ & $-3.8 \pm 23.4$ & $-0.5 \pm 25.9$ \\
\mbox{Lehigh County} & $+0.3 \pm 17.5$ & $-3.1 \pm 16.9$ & $-3.9 \pm 11.8$ & $-0.2 \pm 23.9$ & $-5.0 \pm 20.2$ & $-4.7 \pm 18.4$ & $+0.8 \pm 22.9$ \\
\mbox{Luzerne County} & $+2.5 \pm 28.7$ & $-6.4 \pm 34.6$ & $-7.5 \pm 34.0$ & $-0.6 \pm 40.6$ & $+0.5 \pm 36.2$ & $-1.8 \pm 32.9$ & $+10.6 \pm 44.7$ \\
\mbox{Montgomery County} & $+1.2 \pm 21.4$ & $-3.6 \pm 22.6$ & $-7.4 \pm 18.3$ & $-1.1 \pm 21.2$ & $-3.7 \pm 20.2$ & $-5.1 \pm 21.4$ & $-1.7 \pm 26.2$ \\
\mbox{Montour County} & $+1.1 \pm 28.9$ & $+0.9 \pm 45.2$ & $-2.3 \pm 36.1$ & $+2.6 \pm 59.5$ & $+2.0 \pm 32.9$ & $+0.0 \pm 46.4$ & $+13.3 \pm 52.4$ \\
\mbox{Philadelphia County} & $+0.4 \pm 17.0$ & $-0.6 \pm 16.9$ & $-4.2 \pm 9.3$ & $+0.2 \pm 22.0$ & $-4.6 \pm 18.6$ & $-5.6 \pm 15.8$ & $-3.5 \pm 19.4$ \\
\mbox{Washington County} & $-1.7 \pm 27.7$ & $+0.7 \pm 32.3$ & $-7.2 \pm 25.4$ & $+4.1 \pm 46.6$ & $-4.6 \pm 30.6$ & $-2.1 \pm 32.6$ & $-3.3 \pm 42.8$ \\
\mbox{Westmoreland County} & $-0.1 \pm 31.0$ & $-0.1 \pm 37.8$ & $-3.9 \pm 40.0$ & $+70.0 \pm 584.7$ & $+0.9 \pm 37.1$ & $-0.5 \pm 26.0$ & $+5.4 \pm 53.4$ \\
\mbox{York County} & $-1.0 \pm 20.1$ & $+1.2 \pm 29.9$ & $-4.3 \pm 22.6$ & $+0.8 \pm 45.0$ & $-2.3 \pm 25.2$ & $-4.7 \pm 28.5$ & $-0.1 \pm 28.8$ \\
\bottomrule
\end{tabular}%
}
\end{table*}

\begin{table*}[p]
\centering
\caption{Lead-lag analysis: low-intensity counties (optimal lag $\ell^*$ and peak correlation $\rho^*$).}
\label{tab:leadlag-low-intensity}
{\footnotesize
\setlength{\tabcolsep}{3pt}
\renewcommand{\arraystretch}{0.95}
\resizebox{\textwidth}{!}{%
\begin{tabular}{lrrrrrrrrrrrr}
\toprule
County & \multicolumn{2}{c}{AR(1)} & \multicolumn{2}{c}{ES} & \multicolumn{2}{c}{ARX} & \multicolumn{2}{c}{LLM} & \multicolumn{2}{c}{Hybrid ARX} & \multicolumn{2}{c}{Hybrid LR} \\
\cmidrule(lr){2-3}\cmidrule(lr){4-5}\cmidrule(lr){6-7}\cmidrule(lr){8-9}\cmidrule(lr){10-11}\cmidrule(lr){12-13}
 & $\ell^*$ & $\rho^*$ & $\ell^*$ & $\rho^*$ & $\ell^*$ & $\rho^*$ & $\ell^*$ & $\rho^*$ & $\ell^*$ & $\rho^*$ & $\ell^*$ & $\rho^*$ \\
\midrule
\mbox{Armstrong County} & -1 & 0.787 & -1 & 0.856 & -1 & 0.589 & -1 & 0.907 & -1 & 0.828 & -2 & 0.608 \\
\mbox{Bedford County} & -1 & 0.487 & -1 & 0.814 & -1 & 0.490 & -1 & 0.880 & -1 & 0.487 & -1 & 0.606 \\
\mbox{Carbon County} & -1 & 0.881 & -1 & 0.895 & -1 & 0.698 & -1 & 0.884 & -1 & 0.793 & -1 & 0.612 \\
\mbox{Clarion County} & -1 & 0.822 & -1 & 0.862 & 0 & 0.326 & -1 & 0.912 & -1 & 0.766 & -1 & 0.801 \\
\mbox{Clinton County} & 0 & 0.204 & -1 & 0.885 & 0 & 0.195 & -1 & 0.932 & 0 & 0.200 & -3 & 0.340 \\
\mbox{Columbia County} & -1 & 0.842 & -1 & 0.865 & -1 & 0.774 & -1 & 0.906 & -1 & 0.821 & -2 & 0.581 \\
\mbox{Elk County} & -1 & 0.704 & -1 & 0.865 & -1 & 0.634 & -1 & 0.919 & -1 & 0.828 & -1 & 0.547 \\
\mbox{Fulton County} & -1 & 0.827 & -1 & 0.858 & -1 & 0.768 & -1 & 0.920 & -1 & 0.872 & -2 & 0.452 \\
\mbox{Greene County} & -1 & 0.647 & -1 & 0.818 & -1 & 0.593 & -1 & 0.795 & -1 & 0.610 & -2 & 0.409 \\
\mbox{Huntingdon County} & -1 & 0.637 & -1 & 0.875 & -1 & 0.369 & -1 & 0.925 & -1 & 0.709 & -1 & 0.639 \\
\mbox{Jefferson County} & -1 & 0.904 & -1 & 0.829 & -1 & 0.662 & -1 & 0.908 & -1 & 0.886 & -1 & 0.538 \\
\mbox{Lawrence County} & -1 & 0.858 & -1 & 0.833 & -1 & 0.513 & -1 & 0.800 & -1 & 0.709 & -1 & 0.720 \\
\mbox{McKean County} & -1 & 0.613 & -1 & 0.879 & -1 & 0.503 & -1 & 0.877 & -1 & 0.629 & -2 & 0.504 \\
\mbox{Northumberland County} & -1 & 0.832 & -1 & 0.860 & -1 & 0.762 & -1 & 0.852 & -1 & 0.839 & -1 & 0.668 \\
\mbox{Potter County} & -1 & 0.778 & -1 & 0.868 & -1 & 0.645 & -1 & 0.852 & -1 & 0.744 & -1 & 0.517 \\
\mbox{Susquehanna County} & -1 & 0.835 & -1 & 0.829 & -1 & 0.812 & -1 & 0.899 & -1 & 0.866 & -2 & 0.628 \\
\mbox{Tioga County} & -1 & 0.559 & -1 & 0.872 & -1 & 0.550 & -1 & 0.920 & -1 & 0.588 & -1 & 0.539 \\
\mbox{Warren County} & -1 & 0.656 & -1 & 0.871 & -1 & 0.633 & -1 & 0.913 & -1 & 0.654 & -2 & 0.629 \\
\mbox{Wayne County} & -1 & 0.780 & -1 & 0.818 & 4 & 0.289 & -1 & 0.838 & -1 & 0.870 & -2 & 0.493 \\
\mbox{Wyoming County} & -1 & 0.301 & -1 & 0.763 & -1 & 0.420 & -1 & 0.878 & -1 & 0.442 & -1 & 0.618 \\
\bottomrule
\end{tabular}%
}
}
\end{table*}

\begin{table*}[p]
\centering
\caption{Lead--lag analysis: mid-intensity counties (optimal lag $\ell^*$ and peak correlation $\rho^*$).}
\label{tab:leadlag-mid-intensity}
{\footnotesize
\setlength{\tabcolsep}{3pt}
\renewcommand{\arraystretch}{0.95}
\resizebox{\textwidth}{!}{%
\begin{tabular}{lrrrrrrrrrrrr}
\toprule
County & \multicolumn{2}{c}{AR(1)} & \multicolumn{2}{c}{ES} & \multicolumn{2}{c}{ARX} & \multicolumn{2}{c}{LLM} & \multicolumn{2}{c}{Hybrid ARX} & \multicolumn{2}{c}{Hybrid LR} \\
\cmidrule(lr){2-3}\cmidrule(lr){4-5}\cmidrule(lr){6-7}\cmidrule(lr){8-9}\cmidrule(lr){10-11}\cmidrule(lr){12-13}
 & $\ell^*$ & $\rho^*$ & $\ell^*$ & $\rho^*$ & $\ell^*$ & $\rho^*$ & $\ell^*$ & $\rho^*$ & $\ell^*$ & $\rho^*$ & $\ell^*$ & $\rho^*$ \\
\midrule
\mbox{Adams County} & -1 & 0.747 & -1 & 0.871 & -1 & 0.519 & -1 & 0.876 & -1 & 0.668 & -1 & 0.681 \\
\mbox{Beaver County} & -1 & 0.830 & -1 & 0.894 & -1 & 0.559 & -1 & 0.776 & -1 & 0.810 & -1 & 0.687 \\
\mbox{Bradford County} & -1 & 0.825 & -1 & 0.867 & -1 & 0.570 & -1 & 0.810 & -1 & 0.792 & -1 & 0.654 \\
\mbox{Butler County} & -1 & 0.803 & -1 & 0.877 & -1 & 0.571 & -1 & 0.859 & -1 & 0.872 & -1 & 0.636 \\
\mbox{Cambria County} & -1 & 0.893 & -1 & 0.886 & -1 & 0.646 & -1 & 0.868 & -1 & 0.899 & -1 & 0.647 \\
\mbox{Centre County} & -1 & 0.851 & -1 & 0.848 & -1 & 0.655 & -1 & 0.850 & -1 & 0.862 & -1 & 0.620 \\
\mbox{Clearfield County} & -1 & 0.887 & -1 & 0.864 & -1 & 0.531 & -1 & 0.817 & -1 & 0.852 & -1 & 0.761 \\
\mbox{Crawford County} & -1 & 0.886 & -1 & 0.825 & -1 & 0.698 & -1 & 0.846 & -1 & 0.926 & -1 & 0.717 \\
\mbox{Fayette County} & -1 & 0.829 & -1 & 0.880 & -1 & 0.635 & -1 & 0.833 & -1 & 0.836 & -1 & 0.774 \\
\mbox{Indiana County} & -1 & 0.769 & -1 & 0.851 & -1 & 0.622 & -1 & 0.898 & -1 & 0.816 & -1 & 0.650 \\
\mbox{Lebanon County} & -1 & 0.855 & -1 & 0.822 & -1 & 0.539 & -1 & 0.832 & -1 & 0.844 & -1 & 0.689 \\
\mbox{Lycoming County} & -1 & 0.850 & -1 & 0.890 & 0 & 0.395 & -1 & 0.914 & -1 & 0.743 & -1 & 0.723 \\
\mbox{Mercer County} & -1 & 0.744 & -1 & 0.880 & -1 & 0.428 & -1 & 0.795 & -1 & 0.782 & -1 & 0.679 \\
\mbox{Mifflin County} & -1 & 0.902 & -1 & 0.827 & -1 & 0.643 & -1 & 0.878 & -1 & 0.843 & -1 & 0.624 \\
\mbox{Monroe County} & -1 & 0.884 & -1 & 0.899 & -1 & 0.575 & -1 & 0.820 & -1 & 0.781 & -1 & 0.799 \\
\mbox{Northampton County} & -1 & 0.865 & -1 & 0.905 & -1 & 0.758 & -1 & 0.924 & -1 & 0.917 & -1 & 0.488 \\
\mbox{Schuylkill County} & -1 & 0.887 & -1 & 0.896 & -1 & 0.546 & -1 & 0.861 & -1 & 0.785 & -1 & 0.715 \\
\mbox{Somerset County} & 0 & 0.692 & -1 & 0.862 & -1 & 0.446 & -1 & 0.778 & -1 & 0.766 & -1 & 0.759 \\
\mbox{Union County} & -1 & 0.859 & -1 & 0.852 & -1 & 0.448 & -1 & 0.879 & -1 & 0.705 & -2 & 0.699 \\
\mbox{Venango County} & -1 & 0.840 & -1 & 0.815 & -1 & 0.401 & -1 & 0.874 & -1 & 0.782 & -1 & 0.742 \\
\bottomrule
\end{tabular}%
}
}
\end{table*}

\begin{table*}[p]
\centering
\caption{Lead--lag analysis: high-intensity counties (optimal lag $\ell^*$ and peak correlation $\rho^*$).}
\label{tab:leadlag-high-intensity}
{\footnotesize
\setlength{\tabcolsep}{3pt}
\renewcommand{\arraystretch}{0.95}
\resizebox{\textwidth}{!}{%
\begin{tabular}{lrrrrrrrrrrrr}
\toprule
County & \multicolumn{2}{c}{AR(1)} & \multicolumn{2}{c}{ES} & \multicolumn{2}{c}{ARX} & \multicolumn{2}{c}{LLM} & \multicolumn{2}{c}{Hybrid ARX} & \multicolumn{2}{c}{Hybrid LR} \\
\cmidrule(lr){2-3}\cmidrule(lr){4-5}\cmidrule(lr){6-7}\cmidrule(lr){8-9}\cmidrule(lr){10-11}\cmidrule(lr){12-13}
 & $\ell^*$ & $\rho^*$ & $\ell^*$ & $\rho^*$ & $\ell^*$ & $\rho^*$ & $\ell^*$ & $\rho^*$ & $\ell^*$ & $\rho^*$ & $\ell^*$ & $\rho^*$ \\
\midrule
\mbox{Allegheny County} & -1 & 0.877 & -1 & 0.912 & -1 & 0.552 & -1 & 0.845 & -1 & 0.786 & -1 & 0.751 \\
\mbox{Berks County} & -1 & 0.849 & -1 & 0.865 & -1 & 0.505 & -1 & 0.923 & 0 & 0.800 & -1 & 0.871 \\
\mbox{Blair County} & 0 & 0.775 & -1 & 0.858 & -1 & 0.481 & -1 & 0.598 & -1 & 0.712 & -1 & 0.723 \\
\mbox{Bucks County} & -1 & 0.879 & -1 & 0.922 & -1 & 0.593 & -1 & 0.856 & -1 & 0.858 & -1 & 0.825 \\
\mbox{Chester County} & -1 & 0.915 & -1 & 0.908 & -1 & 0.672 & -1 & 0.845 & -1 & 0.830 & -1 & 0.840 \\
\mbox{Cumberland County} & -1 & 0.846 & -1 & 0.899 & 0 & 0.448 & -1 & 0.834 & -1 & 0.707 & -1 & 0.832 \\
\mbox{Dauphin County} & -1 & 0.863 & -1 & 0.882 & -1 & 0.419 & -1 & 0.856 & -1 & 0.831 & -1 & 0.702 \\
\mbox{Delaware County} & -1 & 0.928 & -1 & 0.939 & -1 & 0.770 & -1 & 0.922 & -1 & 0.896 & -1 & 0.891 \\
\mbox{Erie County} & -1 & 0.857 & -1 & 0.904 & -1 & 0.625 & -1 & 0.827 & -1 & 0.717 & -1 & 0.612 \\
\mbox{Franklin County} & -1 & 0.862 & -1 & 0.863 & -1 & 0.654 & -1 & 0.880 & -1 & 0.825 & -1 & 0.849 \\
\mbox{Lackawanna County} & -1 & 0.899 & -1 & 0.864 & -1 & 0.578 & -1 & 0.819 & -1 & 0.818 & -1 & 0.766 \\
\mbox{Lancaster County} & -1 & 0.893 & -1 & 0.890 & -1 & 0.457 & -1 & 0.771 & -1 & 0.708 & -1 & 0.663 \\
\mbox{Lehigh County} & -1 & 0.872 & -1 & 0.913 & -1 & 0.476 & -1 & 0.867 & -1 & 0.606 & -1 & 0.857 \\
\mbox{Luzerne County} & -1 & 0.868 & -1 & 0.875 & -1 & 0.529 & -1 & 0.821 & -1 & 0.815 & -2 & 0.761 \\
\mbox{Montgomery County} & -1 & 0.935 & -1 & 0.944 & -1 & 0.781 & -1 & 0.951 & -1 & 0.794 & -1 & 0.920 \\
\mbox{Montour County} & -1 & 0.857 & -1 & 0.898 & -1 & 0.595 & -1 & 0.830 & -1 & 0.807 & -1 & 0.672 \\
\mbox{Philadelphia County} & -1 & 0.938 & -1 & 0.955 & 0 & 0.648 & -1 & 0.935 & -1 & 0.897 & -1 & 0.903 \\
\mbox{Washington County} & -1 & 0.876 & -1 & 0.889 & -1 & 0.437 & -1 & 0.770 & -1 & 0.737 & -2 & 0.758 \\
\mbox{Westmoreland County} & -1 & 0.901 & -1 & 0.863 & 0 & 0.430 & -1 & 0.857 & -1 & 0.825 & -1 & 0.715 \\
\mbox{York County} & -2 & 0.675 & -1 & 0.898 & -3 & 0.477 & -1 & 0.841 & -1 & 0.770 & -1 & 0.722 \\
\bottomrule
\end{tabular}%
}
}
\end{table*}

\clearpage

\end{document}